%% file: main.tex
\documentclass[10pt,twocolumn,letterpaper]{article}

\usepackage{cvpr}              
\input{preamble}
\definecolor{cvprblue}{rgb}{0.21,0.49,0.74}
\usepackage[pagebackref,breaklinks,colorlinks,allcolors=cvprblue]{hyperref}
\usepackage{multirow}
\usepackage{multicol}
\usepackage[table]{xcolor}
\definecolor{citecolor}{HTML}{0071BC}
\definecolor{linkcolor}{HTML}{ED1C24}
\definecolor{deemph}{gray}{0.6}
\usepackage{makecell}
\usepackage{enumitem}
\usepackage{float}
\usepackage{circledsteps}
\usepackage{pifont}
\usepackage{wrapfig}
\usepackage{booktabs}
\usepackage{amssymb} 
\definecolor{scoregreen}{RGB}{34, 139, 34}

\def\paperID{*****} 
\def\confName{CVPR}
\def\confYear{2026}

\title{DriveMoE: Mixture-of-Experts for Vision-Language-Action Model in End-to-End Autonomous Driving}

\author{%
  \normalfont
  Zhenjie Yang$^{1,4*}$ \quad
  Yilin Chai$^{1*}$ \quad
  Xiaosong Jia$^{2,3*}$ \quad \\
  Qifeng Li$^{1,4}$ \quad
  Yuqian Shao$^{1,4}$ \quad
  Xuekai Zhu$^{1}$ \quad
  Haisheng Su$^{1}$ \quad
  Junchi Yan$^{1\dagger}$ \\
  \normalsize{$^*$ Equal contributions  \quad $^\dagger$ Correspondence author} \\ \\
  1. Sch. of Computer Science \& Sch. of Artificial Intelligence, Shanghai Jiao Tong University \\
  2. Institute of Trustworthy Embodied AI, Fudan University \\
  3. Shanghai Key Laboratory of Multimodal Embodied AI \quad 4. AnyScale AI \\ \\
  \normalsize{Project Page:~{\url{https://thinklab-sjtu.github.io/DriveMoE/}}}
  }

\begin{document}
\maketitle

\begingroup
\renewcommand\thefootnote{}
\footnotetext{This work was supported by Scientific Research Innovation Capability Support Project for Young Faculty (U40) of the Ministry of Education of China (SRICSPYF-ZY2025019). This work was also in part supported by the Science and Technology Commission of Shanghai Municipality (No. 24511103100) and the New Cornerstone Science Foundation through the XPLORER PRIZE.}
\endgroup

\begin{abstract}
End-to-end autonomous driving (E2E-AD) demands effective processing of multi-view sensor data and robust handling of diverse and complex driving scenarios, particularly rare maneuvers such as aggressive turns. The recent success of the Mixture-of-Experts (MoE) architectures in Large Language Models (LLMs) demonstrates that expert specialization enables strong scalability. In this work, we propose \textbf{DriveMoE}, a novel MoE-based E2E-AD framework, with a \textbf{Scene-Specialized Vision MoE} and a \textbf{Skill-Specialized Action MoE}. First, we introduce \textbf{Drive-$\boldsymbol{\pi}_{\boldsymbol{0}}$}, a Vision-Language-Action (VLA) baseline adapted from Embodied AI for autonomous driving, which serves as the foundation model for DriveMoE. Building on this, we strengthen perception through a carefully designed Vision MoE, where a router adaptively selects context-relevant camera views. This mechanism is inspired by human driving cognition, in which attention is directed to key visual cues rather than to all sensory inputs simultaneously. Beyond perception, we introduce an Action MoE that augments the framework by training a router to activate specialized expert modules tailored to distinct driving behaviors. Within the Action MoE, we implement two distinct styles(Token-level Router and Trajectory-level Router) and extensively explore their applicability in autonomous driving. In Bench2Drive closed-loop evaluations, DriveMoE demonstrates robust performance across diverse driving scenarios, alleviates the mode-averaging effect that limits existing models, and achieves state-of-the-art results with significant improvements over Drive-$\boldsymbol{\pi}_{\boldsymbol{0}}$. 
\end{abstract}

\section{Introduction}
\label{sec:intro}
Modern autonomous driving has made significant progress~\cite{Chitta2023PAMI, hu2023planning, jiang2023vad, su2026drivemamba, su2026egofsd, jia2026guidedvla, jia2025drivevggt, jia2025spatial, zhangtrajtok, han2025percept} with an end-to-end paradigm, which directly maps the raw sensor input into the planning results. This paradigm~\cite{wu2022trajectoryguided, Jaeger2023ICCV} offers several advantages, such as reduced engineering complexity, mitigation of error propagation, and global objective optimization. Despite the encouraging results achieved on various open-loop self-driving benchmarks~\cite{hu2023planning, song2025don, wang2025diffad}, existing end-to-end models still fail to get satisfactory performance in closed-loop settings~\cite{shao2023lmdrive, sima2023drivelm, chen2024asynchronouslargelanguagemodel}. In closed-loop settings, trained driving models can easily encounter out-of-distribution cases~\cite{zhai2023ADMLP, zheng2024genad}, requiring stronger generalization and reasoning abilities. Vision Language Models(VLM) and Vision Language Action Models(VLA) have recently gained much attention due to their strong generalization and transferability across domains~\cite{black2024pi_0, intelligence2025pi_, liu2026palm, xu2025stare, shen2025fine, li2026toward, driess2025knowledge}. 
To enhance generalization and contextual reasoning, recent work~\cite{Yang2023LLM4DriveAS, renz2024carllavavisionlanguagemodels, shao2023lmdrive, sima2023drivelm, mei2024continuously, hamdan2025eta, fu2025orion, jia2026bench2drive} has attempted to introduce VLA into autonomous driving. 

\begin{figure*}[t!]
    \centering
    \includegraphics[width=1\linewidth]{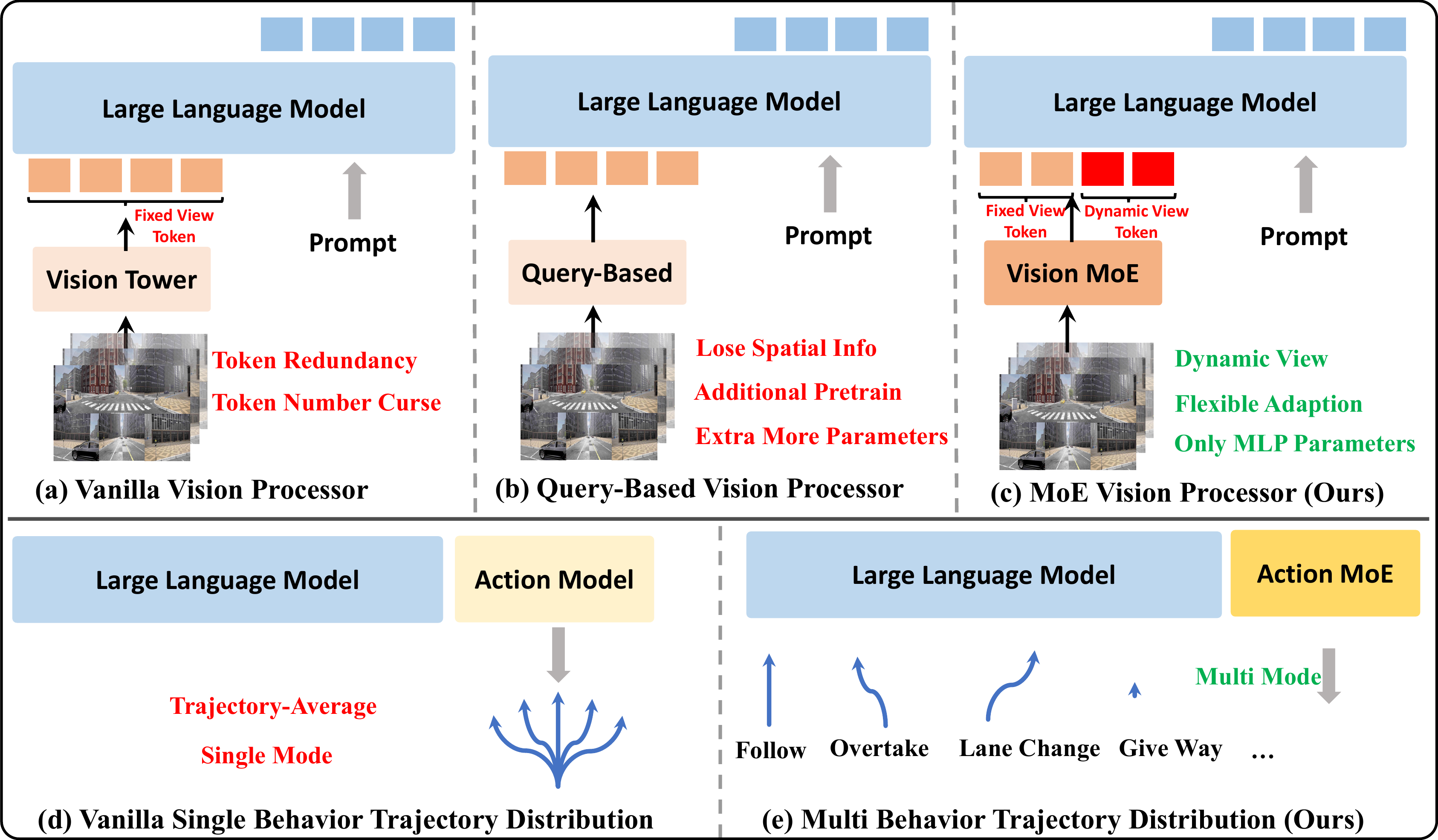}
    \caption{\textbf{Comparison of Different Vision and Action Modeling Strategies in VLA-based End-to-End Driving.} (a) Vanilla visual token encoding~\cite{sima2023drivelm} processes all surround-view images through a vision tower, leading to token redundancy and increased computational cost. (b) Query-based token extraction~\cite{wang2023drivemlm} (e.g., Q-Former~\cite{li2023blip}) selects a subset of visual tokens from each image, but loses spatial structure and requires additional pretraining. (c) Our proposed Scene-Specialized Vision MoE dynamically selects a subset of cameras—typically frontal and a few context-relevant side/rear views, reducing redundancy. (d) Standard action models adopt one policy head to handle all driving scenarios, limiting performance in rare or skill-specific behaviors. (e) Our Skill-Specialized Action MoE, built on a flow-matching planner, activates different experts based on driving intent (e.g., lane following, turning, obstacle avoidance), enabling context-aware and behavior-specialized planning.}
    \label{fig:compare}
\end{figure*}

However, existing VLA approaches still face two major limitations.
Firstly, existing vision processors of VLA introduce information redundancy and significant computational overhead. 
As shown in the upper part of Figure~\ref{fig:compare}, there are two distinct strategies for processing multi-view inputs. 
The first strategy, termed \textbf{vanilla vision processor}~\cite{shao2023lmdrive, sima2023drivelm, chen2024asynchronouslargelanguagemodel, huang2023iddr, huang2026nerf}, processes all available camera views at each timestep without distinction, resulting in a substantial computational burden and redundant visual representations, thereby limiting efficiency and scalability. 
The second strategy, termed \textbf{query-based vision processor}, employs learned queries (e.g., Q-Former modules\cite{li2023blip}) to extract a compact set of visual tokens guided by semantic context. 
However, these learned queries typically lead to the loss of precise geometric and positional information and require substantial additional pre-training efforts~\cite{qi2024gpt4point}. 
Secondly, as shown in the lower part of the Figure~\ref{fig:compare}, current VLA-based frameworks~\cite{renz2024carllavavisionlanguagemodels, shao2023lmdrive} generally employ a single unified policy network designed to handle the full spectrum of driving behaviors. 
Such uniform approaches tend to bias model training towards more frequent scenarios, thereby insufficiently addressing rare but critical driving maneuvers, such as emergency braking or aggressive turning. 
This lack of explicit specialization restricts their effectiveness in dynamically changing and highly context-dependent driving situations. Addressing these two key limitations demands architectural innovations capable of both context-aware dynamic multi-view selection and explicit fine-grained skill specialization. 

Meanwhile, Mixture-of-Experts (MoE) architectures~\cite{cai2025survey, wan2023efficient} have significantly advanced Large Language Models (LLMs)~\cite{zhu2024llama, bai2023qwen, openai2024gpt4technicalreport} by partitioning model capacity into multiple expert modules, scaling to larger model sizes without proportional increases in computational demands. 
Despite their demonstrated success, the extension of MoE principles into the vision and action domains, particularly within autonomous driving, remains largely under-explored. 

Current end-to-end driving models~\cite{jia2025drivetransformer, renz2025simlingo} predominantly rely on predicting trajectory modalities within unified architectures, but lack explicit dynamic expert selection or specialized behavioral adaptation, which limits their scalability. This gap motivates our exploration of MoE-based specialization to enhance both visual perception and decision-making in autonomous driving. Due to limited space, we discuss more details of Related Works in Appendix~\ref{sec:related_work}.

\begin{figure*}[!]
    \centering
    \includegraphics[width=1\linewidth]{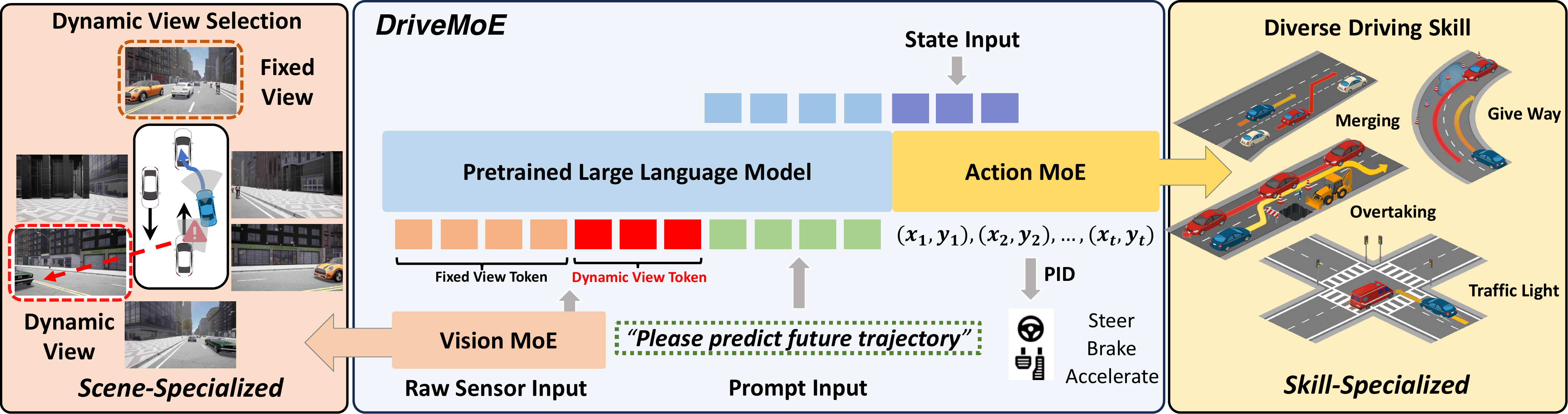}
    \caption{\textbf{Framework of DriveMoE.} Our proposed framework comprises two main Mixture-of-Experts (MoE) modules tailored for end-to-end autonomous driving. The Scene-Specialized Vision MoE dynamically selects relevant camera views based on real-time driving contexts, efficiently reducing visual redundancy. Subsequently, selected views are fused into a unified representation by projector layers. The Skill-Specialized Action MoE, integrated within a flow-matching planner, activates expert controllers specifically optimized for distinct driving behaviors such as merging, overtaking, emergency braking, yielding, and responding to traffic signs. This dual MoE structure enhances computational efficiency, adaptability, and robustness to rare, safety-critical driving scenarios.}
    \label{fig:drivemoe}
\end{figure*}

To address these challenges, we propose \textbf{\textit{DriveMoE}}, a novel framework built upon our proposed Drive-$\boldsymbol{\pi}_{\boldsymbol{0}}$, a Vision-Language-Action (VLA) foundation model extended from the embodied AI model $\boldsymbol{\pi}_{\boldsymbol{0}}$~\cite{black2024pi_0, intelligence2025pi_}. \textbf{\textit{DriveMoE}} introduces both a \textbf{\textit{Scene-Specialized Vision MoE}} and a \textbf{\textit{Skill-Specialized Action MoE}}, carefully designed for end-to-end autonomous driving scenarios. DriveMoE dynamically selects contextually relevant camera views and activates skill-specific experts for specialized planning. The Vision MoE employs a learned router to dynamically prioritize camera views aligned with the immediate driving context, integrating projector layers that fuse these selected views into a cohesive visual representation. This approach mirrors human attentional strategies, allowing efficient processing of only critical visual inputs. Concurrently, the Action MoE leverages another routing mechanism to engage distinct experts within a flow-matching planning architecture~\cite{lipman2022flow}, with each expert dedicated to handling specialized behaviors such as lane following, obstacle avoidance, or aggressive maneuvers. By introducing context-driven dynamic expert selection across both perception and planning modules, DriveMoE ensures efficient resource utilization and robust specialization, significantly improving handling of rare, complex, and long-tail driving behaviors. \textbf{The contributions are as follows:}

\begin{itemize}[noitemsep,topsep=0pt,leftmargin=*,itemsep=2pt]
    \item We extend the VLA foundation model \textbf{\textit{$\boldsymbol{\pi}_{\boldsymbol{0}}$}}, originally designed for embodied AI, into the autonomous driving, developing \textbf{\textit{Drive-$\boldsymbol{\pi}_{\boldsymbol{0}}$}} as a unified framework for visual perception, contextual understanding, and action planning.
    \item Recognizing differences between embodied AI and autonomous driving, we propose \textbf{\textit{DriveMoE}}, the first framework integrating Mixture-of-Experts (MoE) into perception and decision-making to address inefficiencies in multi-view processing and diverse driving behaviors.
    \item We design a \textbf{\textit{Scene-specialized Vision MoE}} for dynamic camera view selection and a \textbf{\textit{Skill-specialized Action MoE}} for behavior-specific planning, addressing challenges of multi-view redundancy and skill specialization.
    \item We demonstrate that DriveMoE achieves state-of-the-art (SOTA) performance on the Bench2Drive closed-loop simulation benchmark, significantly improving robustness to rare driving behaviors.
\end{itemize}

\begin{figure*}[t!]
    \centering
    \includegraphics[width=1\linewidth]{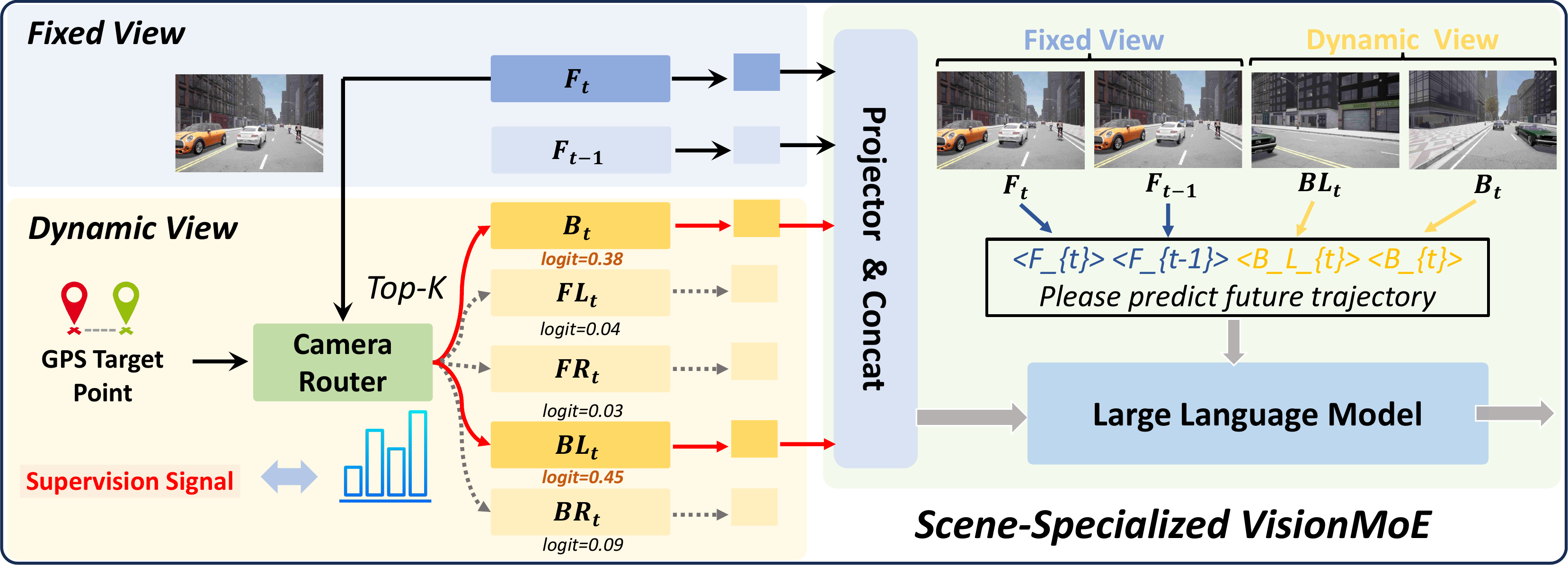}
    \caption{\textbf{The Scene-Specialized Vision Mixture-of-Experts.}}
    \label{fig:visionmoe}
\end{figure*}  

\section{Method}
\label{sec:method}
\subsection{Preliminary: Drive-$\boldsymbol{\pi}_{\boldsymbol{0}}$ Baseline}
We first establish a strong baseline, Drive-$\boldsymbol{\pi}_{\boldsymbol{0}}$, which builds upon the recently proposed $\boldsymbol{\pi}_{\boldsymbol{0}}$~\cite{black2024pi_0} Vision-Language-Action (VLA) framework from embodied AI, and extends it to the domain of end-to-end autonomous driving. As shown in Figure~\ref{fig:drivemoe}, specifically, the input to Drive-$\pi_0$ includes: (i) two consecutive front-view images from onboard multi-camera sensors; (ii) a fixed text prompt (e.g., "Please predict future trajectory"); and (iii) the current vehicle state (e.g., speed, yaw rate, and past trajectory). The network design follows $\boldsymbol{\pi}_{\boldsymbol{0}}$ framework with pre-trained Paligemma VLM~\cite{beyer2024paligemma} as the backbone and a flow-matching-based action module for planned future trajectory generation. 

\subsection{Motivation: From Drive-$\boldsymbol{\pi}_{\boldsymbol{0}}$ to DriveMoE}
With Drive-$\boldsymbol{\pi}_{\boldsymbol{0}}$ as the baseline, we identify two major challenges: (i) adopting VLM to process spatial-temporal surround-view video tokens poses significant challenges to computational resource; (ii) driving performance for rare and difficult scenarios are deficient, even if there is similar data for training. It might be related to the interference effect of different behaviors, as mentioned in the $\boldsymbol{\pi}_{\boldsymbol{0}}$ paper~\cite{black2024pi_0}. Inspired by the recent success of Mixture of Experts (MoE) in VLM field~\cite{jiang2024mixtral,deepseekai2024deepseekv3technicalreport}, we introduce \textbf{DriveMoE}, which extends Drive-$\boldsymbol{\pi}_{\boldsymbol{0}}$ by adding two Mixture-of-Experts (MoE) modules to tackle the aforementioned challenges: (i) We propose a Scene-Specialized Vision MoE that dynamically selects the most relevant camera views based on the current driving context, effectively reducing redundant visual tokens. (ii) We incorporate a Skill-Specialized Action MoE within a flow-matching transformer to generate more accurate future trajectory distributions tailored to diverse driving skills. Figure~\ref{fig:drivemoe} illustrates the DriveMoE architecture. 

\subsection{Scene-Specialized Vision MoE}
Typical Vision-Language-Action Models (VLAs)~\cite{kim2024openvla, black2024pi_0} usually handle only a single or a few images at a time, whereas autonomous driving must handle multi-view, multi-timestep visual inputs. Concatenating all camera frames into a transformer leads to a visual token bottleneck – an explosion in sequence length that drastically slows training and inference and hampers convergence. Among existing works, \cite{shao2023lmdrive, sima2023drivelm} adopts a vanilla vision processor to directly handle all visual tokens, while query-based compression modules (e.g. Q-Former~\cite{li2023blip}) reduce token count but sacrifice spatial structure, often treating images as a “bag of patches” without fine spatial correspondence~\cite{qi2024gpt4point}.

In this work, we seek a simple and efficient approach that reduces the token load without losing the rich spatial context crucial for driving. \textbf{\textit{Inspired by human drivers—who naturally prioritize specific visual information based on driving context}}—we propose a \textbf{\textit{Scene-Specialized Vision Mixture-of-Experts (Vision MoE)}} module. Specifically, as shown in Figure~\ref{fig:visionmoe}, our Vision MoE dynamically selects a subset of the most relevant camera views according to the current driving situation and future goal waypoint provided by the route planner. Unlike token-level annotations (which are impractical and costly), camera-view annotations are straightforward and inexpensive, allowing human priors to be integrated effectively. This dynamic attention strategy significantly reduces the number of visual tokens processed per timestep, greatly improving computational efficiency and decision accuracy.

Formally, we define the image from camera view $v$ at timestep $t$ as $\boldsymbol{I}_t^v$, where $v \in \{1,2,\dots,N\}$ for $N$ available camera views. In particular, the front-view image at timestep $t$ is denoted by $\boldsymbol{I}^{\text{front}}_{t}$. We introduce a lightweight vision router module $\boldsymbol{R}_{\text{vision}}$, which takes as inputs the front-view embedding $\boldsymbol{e}^{\text{front}}_{t}$ and the future goal waypoint $\boldsymbol{g}_t$, computing a probability distribution $\boldsymbol{p}_t \in \mathbb{R}^{N}$ across all views:
\begin{equation}
    \boldsymbol{p}_t = \text{S(o}\left(\boldsymbol{R}_{\text{vision}}\left(\boldsymbol{e}^{\text{front}}_{t}, \boldsymbol{g}_t\right)\right),
\end{equation}
where each element $p_t^v$ indicates the selection probability of camera view $v$ at timestep $t$. Notably, this routing happens before the expensive backbone computation, so views not selected can be skipped entirely to save compute. Thus, we obtain the input for VLM: \textlangle fixed\_view\textrangle, \textlangle fixed\_view\textrangle, \textlangle dynamic\_view\textrangle, \textlangle dynamic\_view\textrangle, \textlangle text\textrangle, \textlangle text\textrangle.

\begin{figure*}[!t]
    \centering
    \includegraphics[width=0.9\linewidth]{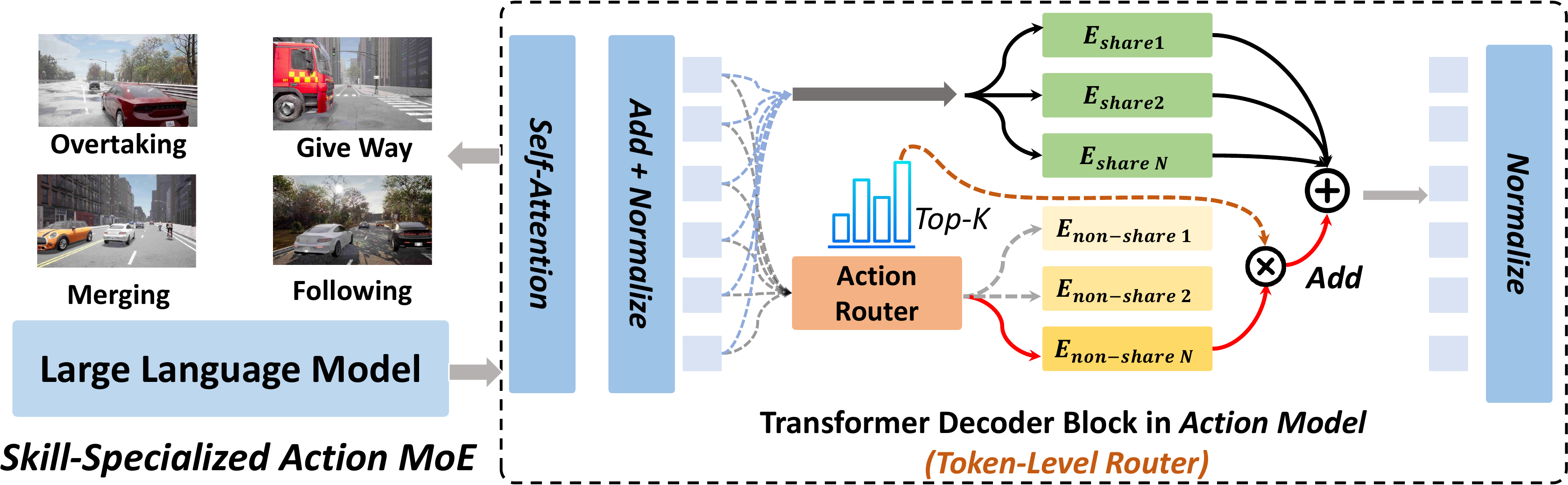}
    \caption{\textbf{Token-Level Skill-Specialized Action Mixture-of-Experts.}}
    \label{fig:actionmoe_token}
\vspace{-5pt}
\end{figure*}

We further incorporate learnable positional embeddings ($\textbf{\text{PE}}$) that are unique to each camera viewpoint into their corresponding vision tokens to preserve spatial and positional relation across different camera views. The label for selection of views  is annotated by manually designed filters based on future trajectories, bounding box, and maps, detailed in Appendix~\ref{sup:moe_label}. With the annotated binary camera-view selection labels $\boldsymbol{y_{t}} \in \{0,1\}$ , the vision router is trained using the cross-entropy loss:
\begin{equation}
    \mathcal{L}_{\text{Vision-Router}} = -\lambda_0\sum_{v=1}^{N}\boldsymbol{y_{t}^v}\log(\boldsymbol{p}_t^v),
\end{equation}
which explicitly encourages the model to proactively select informative camera views relevant for decision-making. $\lambda_0$ represents the loss weight of vision router.

\subsection{Skill-Specialized Action MoE}
Human drivers fluidly transition among different driving skills—such as smoothly cruising down a highway, carefully merging into traffic, swiftly overtaking slower vehicles, or urgently braking in response to sudden obstacles. Each of these driving skills is associated with distinct behavioral patterns and trajectory characteristics. Although the original flow-matching decoder of $\boldsymbol{\pi}_{\boldsymbol{0}}$ could already generate diverse trajectories, employing one single model inevitably averages across these diverse behaviors~\cite{black2024pi_0},  making the model fail to accurately generate rare yet safety-critical maneuvers.

\begin{figure*}[!t]
    \centering
    \includegraphics[width=0.9\linewidth]{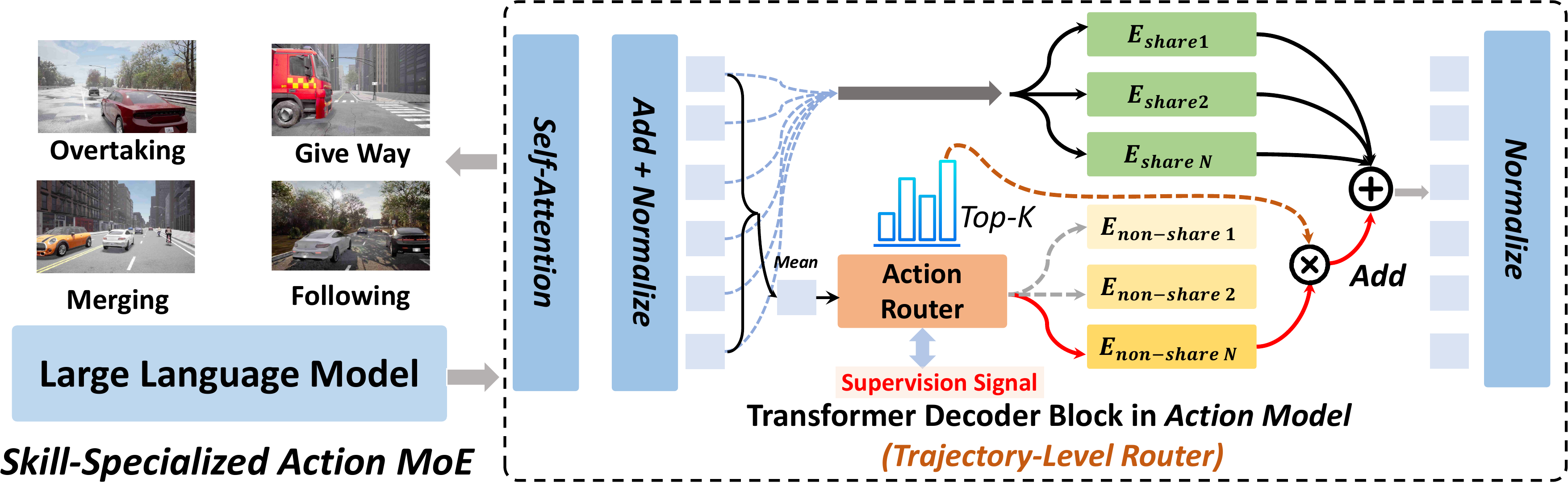}
    \caption{\textbf{Trajectory-Level Skill-Specialized Action Mixture-of-Experts.}}
    \label{fig:actionmoe_traj}
\end{figure*}

To address these issues, inspired by human intuition—where drivers naturally select the appropriate driving skill based on the current context, we propose a Skill-Specialized Action MoE architecture built on the original flow-matching trajectory transformer. The details about flow matching trajectory loss refer to Appendix~\ref{sec:flow_loss}. The central idea is to decompose the policy’s representation of behaviors by replacing each dense feed-forward network (FFN) in the decoder with a Mixture-of-Experts (MoE) layer containing multiple skill-specific experts. Essentially, each decoder layer is no longer a single monolithic mapping, but an ensemble of $K$ expert FFNs each intended to specialize in a subset of driving skills. By conditionally routing each input through a small subset of these experts, the model isolates distinct behavior modes instead of forcing them into a single decoder stream. This design prevents the averaging effect observed in one single model and thereby allocates dedicated model capacity to rare maneuvers. The result is a policy network that preserves the multimodality of the trajectory data, modeling both frequent and infrequent behaviors with appropriate specialization. As illustrated in Figure~\ref{fig:actionmoe_token} and Figure~\ref{fig:actionmoe_traj}, we explore two styles of MoE design: token-level (similar to DeepSeek-MoE) and trajectory-level.

\subsubsection{Token-level ActionMoE}
In Token-level ActionMoE, each token (corresponding to a specific timestep in the trajectory) independently selects its expert(s) based on its local hidden representation. This strategy allows different experts to specialize in modeling short-horizon temporal dependencies—e.g., acceleration, braking, or turning subtasks—within a single trajectory. 

Formally, consider a Transformer decoder layer $\ell$ with input hidden state $\mathbf{h}^{(\ell-1)} \in \mathbb{R}^d$. We introduce $\mathbf{K}$ shared expert models ${{E_{\text{share}}}_{1}^{(\ell)}}$, ${{E_{\text{share}}}_{2}^{(\ell)}}$, $\dots$, ${{E_{\text{share}}}_{K}^{(\ell)}}$ and $\mathbf{M}$ non-shared expert models ${{E_{\text{non-share}}}_{1}^{(\ell)}}$, ${{E_{\text{non-share}}}_{2}^{(\ell)}}$, $\dots$, ${{E_{\text{non-share}}}_{M}^{(\ell)}}$ in this layer, each an independent FFN with its own parameters. Each expert processes the input to produce an output $\boldsymbol{y}^{(\ell)} = E^{(\ell)}(\mathbf{h}^{(\ell-1)})$. In parallel, an action router $\boldsymbol{R}_{\text{action}}$ computes a set of non-shared routing logits ${r_{1}^{(\ell)}, \dots, r_{K}^{(\ell)}}$ based on the same input. We then convert these logits into a probability distribution over experts via a softmax:
\begin{equation}
    \boldsymbol{r}_k^{(\ell-1)} = \text{Softmax}(\boldsymbol{R}_{\text{action}}(\mathbf{h}^{(\ell-1)})), \quad k \in \{1,2,\dots, \boldsymbol{K}\}.
\end{equation}
It is worth noting that $\mathbf{h}$ is calculated in the dimension of token num. The updated feature combines the outputs of individual experts weighted by the router’s confidence:
\begin{equation}
    \boldsymbol{h}^{(\ell)} = \sum_{k=1}^{K}\boldsymbol{r}_k^{(\ell-1)}\boldsymbol{y}_k^{(\ell-1)} + \sum_{m=1}^{M}\boldsymbol{y}_m^{(\ell-1)}
\end{equation}
In practice, we use a sparse activation mechanism~\cite{jiang2024mixtral} to select only a few experts with the highest ranking for calculation (only activate the Top-1 or Top-2 experts), thereby reducing the amount of calculation, preventing mutual interference between experts, and strengthening the degree of expert skill specialization. This sparse routing mechanism is consistent with the mechanism we use in the Vision MoE module, ensuring that each expert clearly focuses on a specific behavior mode.

\begin{table*}[tb!]
\caption{\textbf{Performance on Bench2Drive Multi-Ability Benchmark}. *: expert feature distillation. \label{tab:ability}}
\centering
 \resizebox{0.8\textwidth}{!}{
\begin{tabular}{l|c|cccccc}
\toprule
\multirow{2}{*}{\textbf{Method}} &
  \multirow{2}{*}{\textbf{Venue}} &
  \multicolumn{6}{c}{\textbf{Ability} (\%) $\uparrow$} \\ \cline{3-8}
 &
   &
  \multicolumn{1}{c|}{Merging} &
  \multicolumn{1}{c|}{Overtaking} &
  \multicolumn{1}{c|}{Emergency Brake} &
  \multicolumn{1}{c|}{Give Way} &
  \multicolumn{1}{c|}{Traffic Sign} &
  \textbf{Mean} \\ \hline
TCP-traj*~\cite{wu2022trajectoryguided} &
  NeurIPS 2022 &
  \multicolumn{1}{c|}{8.89} &
  \multicolumn{1}{c|}{24.29} &
  \multicolumn{1}{c|}{51.67} &
  \multicolumn{1}{c|}{40.00} &
  \multicolumn{1}{c|}{46.28} &
  34.22 \\ 
AD-MLP~\cite{zhai2023rethinking} &
  Arxiv 2023 &
  \multicolumn{1}{c|}{0.00} &
  \multicolumn{1}{c|}{0.00} &
  \multicolumn{1}{c|}{0.00} &
  \multicolumn{1}{c|}{0.00} &
  \multicolumn{1}{c|}{4.35} &
  0.87 \\
UniAD-Base~\cite{hu2023planning} &
  CVPR 2023 &
  \multicolumn{1}{c|}{14.10} &
  \multicolumn{1}{c|}{17.78} &
  \multicolumn{1}{c|}{21.67} &
  \multicolumn{1}{c|}{10.00} &
  \multicolumn{1}{c|}{14.21} &
  15.55 \\
ThinkTwice*~\cite{jia2023thinktwice} &
  CVPR 2023 &
  \multicolumn{1}{c|}{27.38} &
  \multicolumn{1}{c|}{18.42} &
  \multicolumn{1}{c|}{35.82} &
  \multicolumn{1}{c|}{\textbf{50.00}} &
  \multicolumn{1}{c|}{54.23} &
  37.17 \\
VAD~\cite{jiang2023vad} &
  ICCV 2023 &
  \multicolumn{1}{c|}{8.11} &
  \multicolumn{1}{c|}{24.44} &
  \multicolumn{1}{c|}{18.64} &
  \multicolumn{1}{c|}{20.00} &
  \multicolumn{1}{c|}{19.15} &
  18.07 \\
DriveAdapter*~\cite{jia2023driveadapter} &
  ICCV 2023 &
  \multicolumn{1}{c|}{28.82} &
  \multicolumn{1}{c|}{26.38} &
  \multicolumn{1}{c|}{48.76} &
  \multicolumn{1}{c|}{\textbf{50.00}} &
  \multicolumn{1}{c|}{56.43} &
  42.08 \\ 
DriveTrans~\cite{jia2025drivetransformer} &
  ICLR 2025 &
  \multicolumn{1}{c|}{17.57} &
  \multicolumn{1}{c|}{35.00} &
  \multicolumn{1}{c|}{48.36} &
  \multicolumn{1}{c|}{40.00} &
  \multicolumn{1}{c|}{52.10} &
  38.60 \\
DiffAD~\cite{wang2025diffad} &
  Arxiv 2025  &
  \multicolumn{1}{c|}{30.00} &
  \multicolumn{1}{c|}{35.55} &
  \multicolumn{1}{c|}{46.66} &
  \multicolumn{1}{c|}{40.00} &
  \multicolumn{1}{c|}{46.32} &
  38.79  \\
  \hline
\rowcolor{gray!20} Drive-$\boldsymbol{\pi}_{\boldsymbol{0}}$ (Ours) &
   &
  \multicolumn{1}{c|}{26.25} &
  \multicolumn{1}{c|}{26.67} &
  \multicolumn{1}{c|}{45.00} &
  \multicolumn{1}{c|}{30.00} &
  \multicolumn{1}{c|}{38.95} &
  33.37 \\
  \rowcolor{gray!20} DriveMoE (Token-Level) &
  Ours &
  \multicolumn{1}{c|}{\textbf{28.75}} &
  \multicolumn{1}{c|}{\textbf{31.11}} &
  \multicolumn{1}{c|}{\textbf{51.67}} &
  \multicolumn{1}{c|}{40.00} &
  \multicolumn{1}{c|}{\textbf{52.63}} &
  \textbf{40.83} \\ 
 \rowcolor{gray!20} DriveMoE (Traj-Level) &
   &
  \multicolumn{1}{c|}{\textbf{34.67}} &
  \multicolumn{1}{c|}{\textbf{40.00}} &
  \multicolumn{1}{c|}{\textbf{65.45}} &
  \multicolumn{1}{c|}{40.00} &
  \multicolumn{1}{c|}{\textbf{59.44}} &
  \textbf{47.91} \\\bottomrule
\end{tabular}
}
\vspace{-10pt}
\end{table*}

\subsubsection{Trajectory-level ActionMoE}
In Trajectory-level ActionMoE, instead of routing per token, the router assigns experts at the trajectory level. 

As shown in Figure~\ref{fig:actionmoe_traj}, unlike the token-level Action MoE, this variant averages the entire token sequence before feeding it into the router, effectively performing expert selection along the batch (trajectory) dimension rather than the token dimension. Each trajectory is treated as a single entity representing a specific scenario or driving skill.

Moreover, to explicitly guide our model toward meaningful skill specialization—mirroring structured and intuitive human-defined skill categories—we utilize driving skill labels $y_k\in\{1,\dots,K\}$, annotated based on scenarios, and train the skill router via a cross-entropy loss as well:
\begin{equation}
    \mathcal{L}_{\text{Action-Router}} = -\boldsymbol{y_{k}}\log(\boldsymbol{r}_k)
\end{equation}
Additionally, we optimize the entire Action MoE module using a flow-matching trajectory loss $\mathcal{L}_{\text{FM}}$ to ensure accurate trajectory predictions:
\begin{equation}
    \mathcal{L}_{\text{Action}} = \lambda_1\mathcal{L}_{\text{FM}} + \lambda_2\mathcal{L}_{\text{Action-Router}}
\end{equation}
where $\lambda_1$ represents loss weight of flow matching policy, $\lambda_2$ represents loss weight of action router. We introduce noise into the action router following~\cite{dai2024deepseekmoe}, which increases randomness and encourages exploration, effectively mitigating the risk of expert collapse. 

\subsection{Two Stage Training: From Teacher-Forcing to Adaptive Training}
DriveMoE loads the Paligemma VLM pretrained weights~\cite{beyer2024paligemma} and we finetune it in the autonomous driving scene via a  \textbf{\textit{two-stage training}} procedure. In the first stage, both vision and action MoEs only select ground-truth experts while the router is trained jointly, which significantly stabilize the training. In the second stage, we transition to select experts based on the outputs of Vision and Action MoE routers, removing reliance on GT annotation about experts. It develops robustness against potential mistakes or inaccuracies made by routers, thereby enhancing the overall model's generalization capability under realistic inference conditions.



\begin{table*}[t]
\centering
\caption{\textbf{Results on the Bench2Drive Benchmark(\textbf{Closed-Loop} and \textbf{Open-Loop})}. * denotes expert feature distillation.
\label{tab:drivescore}}
\resizebox{0.95\textwidth}{!}{
\begin{tabular}{l|c|cccc|c}
\cline{1-7}
\toprule
\multirow{2}{*}{\textbf{Method}}                                & \multirow{2}{*}{\textbf{Venue}} & \multicolumn{4}{c|}{\textbf{Closed-loop Metric}}                                                                                                                              & \textbf{Open-loop Metric}    \\ \cline{3-7}
                                                                &                        & DS $\uparrow$ & SR(\%) $\uparrow$ & Efficiency $\uparrow$ &Comfort $\uparrow$ & Avg. L2$\downarrow$    \\ \hline
TCP-traj*~\cite{wu2022trajectoryguided}   & NeurIPS 2022           & 59.90                                 & 30.00                                   & 76.54                                         & 18.08                                      & 1.70                  \\
AD-MLP~\cite{zhai2023rethinking}          & Arxiv 2023             & 18.05                                 & 0.00                                    & 48.45                                         & 22.63                                      & 3.64                  \\
VAD~\cite{jiang2023vad}                   & ICCV 2023              & 42.35                                 & 15.00                                   & 157.94                                        & 46.01                             & 0.91                  \\
UniAD-Base~\cite{hu2023planning}          & CVPR 2023              & 45.81                                 & 16.36                                   & 129.21                                        & 43.58                                      & 0.73                  \\
ThinkTwice*~\cite{jia2023thinktwice}      & CVPR 2023              & 62.44                                 & 31.23                                   & 69.33                                         & 16.22                                      & 0.95                  \\
DriveAdapter*~\cite{jia2023driveadapter}  & ICCV 2023              & 64.22                                 & 33.08                                   & 70.22                                         & 16.01                                      & 1.01                  \\
GenAD~\cite{zheng2024genad}               & ECCV 2024              & 44.81                                 & 15.90                                   & -                                             & -                                          & \multicolumn{1}{c}{-} \\
DriveTrans~\cite{jia2025drivetransformer} & ICLR 2025              & 63.46                                 & 35.01                                   & 100.64                                        & 20.78                                      & \textbf{0.62}                  \\
MomAD~\cite{song2025don}                  & CVPR 2025              & 44.54                                 & 16.71                                   & 170.21                                        & \textbf{48.63}                                      & 0.82                  \\
WoTE~\cite{li2025end}                     & ICCV 2025             & 61.71                                & 31.36                                   & -                                             & -                                          & \multicolumn{1}{c}{-} \\
DriveMamba-L~\cite{su2026drivemamba} & ICLR 2026 & 66.82 & 37.73 & 152.91 & 18.77 & 0.70 \\
DiffAD~\cite{wang2025diffad}              & Arxiv 2025             & 67.92                                 & 38.64                                   & -                                             & -                                          & 1.55                  \\  
Raw2Drive~\cite{yang2025raw2drive} & NeurIPS 2025 & 71.36 & 50.24 & 214.17 & 22.42 & - \\

\hline
\rowcolor{gray!20} Drive-$\boldsymbol{\pi}_{\boldsymbol{0}}$             &                       & 55.85                          & 30.00                  & 173.63          & 35.70                                       &  1.13                 \\
\rowcolor{gray!20} DriveMoE (Token-Level)             & Ours                      & 66.94                         & 35.45                            & 158.80                                  & 6.86                                        & 0.96                  \\ 
\rowcolor{gray!20} DriveMoE (Traj-Level)             &                       & \textbf{74.22}                          & \textbf{48.64}                            & \textbf{175.96}                                  & 15.31                                        & 1.01                \\
\bottomrule
\end{tabular}
}
\end{table*}

\section{Experiments}
\label{sec:experiments}
\subsection{Datasets \& Benchmark \& Metric}
We employ the CARLA simulator~\cite{dosovitskiy2017carla} (version 0.9.15.1) for closed-loop driving performance evaluation, and adopt the latest public closed-loop evaluation benchmark, Bench2Drive~\cite{jia2024bench} which includes 220 routes with one challenging corner case per route for analysis of different driving abilities. It provides an official training set, where we use the base set (1000 clips, 950 training, 50 test/validation) for fair comparison with all the other baselines.

\begin{table*}[t]
\centering
\caption{\textbf{Ablation study on Vision MoE.}  Compare different camera view combinations and supervision signals. $F$, $FL$, $FR$, and $B$ indicate the front, front-left, front-right, and back views, respectively, while $BL$ and $BR$ represent the back-left and back-right views. \textbf{Fixed View} means selecting a specific view. \textbf{Dynamic View} refers to the camera view dynamically selected by the vision router as the top-1 relevant view according to scene context. Exp 1 denotes our baseline \textbf{Drive-$\boldsymbol{\pi}_{\boldsymbol{0}}$}, which models surrounding agents’ velocities from two consecutive front-view images, and Exp 9 denotes \textbf{DriveMoE}, which adds a dynamically selected view with explicit supervision to enhance perception learning. Memory is evaluated at \textbf{batch size=1}. \textbf{All experiments use the previous-frame front view by default.}}
\resizebox{0.8\linewidth}{!}{
\begin{tabular}{c|cccccc|c|c|c|c|c|c}
\toprule
\textbf{Exp} & $I_F$ & $I_{FL}$ & $I_{FR}$ & $I_B$ & $I_{BL}$ & $I_{BR}$ & \textbf{View} & \textbf{Supervision} & \textbf{DS$\uparrow$} & \textbf{SR(\%)$\uparrow$} & \textbf{Latency$\downarrow$} & \textbf{Memory(MB)} \\
\midrule
1 & $\checkmark$ & $\times$ & $\times$ & $\times$ & $\times$ & $\times$ & Fixed&-& 55.85 & 30.00 & 100ms & 4100 \\
2 & $\checkmark$ & $\checkmark$ & $\times$ & $\times$ & $\times$ & $\times$ & Fixed&-& 62.38 & 33.64 & 260ms & 5100 \\
3 & $\checkmark$ & $\times$ & $\checkmark$ & $\times$ & $\times$ & $\times$ & Fixed&-& 61.52 & 32.73 & 260ms & 5100 \\
4 & $\checkmark$ & $\times$ & $\times$ & $\checkmark$ & $\times$ & $\times$ & Fixed&-& 63.26 & 31.82 & 260ms & 5100 \\
5 & $\checkmark$ & $\checkmark$ & $\checkmark$ & $\times$ & $\times$ & $\times$ & Fixed&-& 64.92 & 33.64 & 400ms & 7400 \\
6 & $\checkmark$ & $\checkmark$ & $\checkmark$ & $\checkmark$ & $\times$ & $\times$ & Fixed&-& 64.18 & 33.64 & 550ms & 9600 \\
7 & $\checkmark$ & $\checkmark$ & $\checkmark$ & $\checkmark$ & $\checkmark$ & $\checkmark$ & Fixed &-& 62.27 & 31.36 & 700ms & 11800 \\
8 & $\checkmark$ &-&-&-&-&-&Dynamic&$\times$& 69.71 & 44.09 & 260ms & 5100 \\
\rowcolor{gray!20} 9 & $\checkmark$ &-&-&-&-&-&Dynamic&$\checkmark$& 74.22& 48.64 & 260ms & 5100 \\
\bottomrule
\end{tabular}}
\label{tab:ablation_vison_moe}
\end{table*}

\subsection{Comparison with State-of-the-Art Methods}
As shown in Table~\ref{tab:drivescore}, our proposed method achieves state-of-the-art (SOTA) performance in terms of both driving score and success rate on the Bench2Drive closed-loop benchmark. Specifically, compared to the baseline Drive-$\boldsymbol{\pi}_{\boldsymbol{0}}$, our method improves the driving score by 22.8\% and the success rate by 62.1\%. On the open-loop metric, our method achieves the lowest L2 error. We observe that diffusion policy-based trajectory prediction significantly reduces L2 errors compared to traditional methods. However, as highlighted in prior studies such as AD-MLP~\cite{zhai2023ADMLP}, TransFuser++~\cite{Jaeger2023ICCV}, and Bench2Drive~\cite{jia2024bench}, open-loop metrics mainly serve as indicators of model convergence, whereas closed-loop metrics provide a more reliable assessment of true driving performance. Moreover, in the multi-dimensional capability evaluation, as shown in Table~\ref{tab:ability}, our method obtains state-of-the-art results across five key capabilities and their overall average.

We use the official 220 routes and official metrics of Bench2Drive for evaluation. The \textbf{Driving Score (DS)} is defined as the product of Route Completion and Infraction Score, capturing both task completion and rule adherence. The \textbf{Success Rate (SR)} measures the percentage of routes completed successfully within the allocated time and without committing any traffic violations. \textbf{Efficiency} quantifies the vehicle’s velocity relative to surrounding traffic, encouraging progressiveness without aggression. \textbf{Comfort} reflects the smoothness of the driving trajectory. Meanwhile, Bench2Drive evaluates driving capabilities across multiple critical dimensions, including tasks such as \textbf{Merging}, \textbf{Overtaking}, \textbf{Emergency Braking}, \textbf{Yielding}, and \textbf{Traffic Signs}. All results are averaged over three runs.

\subsection{Ablation Study}
\noindent
\textbf{Vision MoE.} 
As shown in Table~\ref{tab:ablation_vison_moe}, we investigate the contribution of camera view selection and supervision signals within our Vision MoE module. The baseline (Exp 1, Drive-$\boldsymbol{\pi}_{\boldsymbol{0}}$) utilizes two consecutive front-view images primarily to estimate velocities of surrounding agents. Adding a third fixed view such as the back view (Exp 4), front-left view (Exp 2), or front-right view (Exp 3) provides additional spatial context, yielding moderate improvements. According to Exp 2-4, adding individual additional views provides more complementary perceptual information, leading to improved performance. However, when multiple views are added simultaneously (as in Exp 5-7), the number of visual tokens increases significantly, which introduces greater training complexity and convergence difficulty. As a result, model performance degrades, and the latency increases. 

By introducing dynamically selected views without supervision (Exp 8), the driving score and success rate significantly improve. Ultimately, incorporating explicit supervision signals (Exp 9, DriveMoE) for the dynamic view selection further enhances both driving score and success rate, demonstrating the effectiveness of our Vision MoE module in leveraging dynamic and supervised multi-view perception. Table~\ref{tab:router_acc} shows the accuracy of the vision and action routers on the test set under open-loop evaluation. 

\noindent
\textbf{Token vs Trajectory Level Action MoE.} As shown in Table~\ref{tab:style_action}, we compare the token-level and trajectory-level routers, and observe that the trajectory-level router consistently outperforms the token-level counterpart. So We use \textbf{\textit{Trajectory Level Action MoE as our default Action MoE.}}

\noindent
\textbf{Action MoE.} For clarity, the term Action MoE in the following sections specifically refers to the Trajectory-Level Action MoE. We investigate the impact of the number of non-shared experts within our Action MoE, as shown in Table~\ref{tab:ablation_action_moe}. Specifically, Exp 1 corresponds to the original five skills defined by Bench2Drive~\cite{jia2024bench}, while Exp 2 introduces an additional expert for the classic \textit{ParkingExits} scenario, resulting in improved performance. To further analyze the effect of expert specialization, we conducted additional experiments: Exp 2–3 compare action supervision strategies, where the action supervision yields noticeable improvement; Exp 4-5 introduces additional shared experts, whose performance remains comparable to the Exp 2; Moreover, Exp 6 adds experts targeting several challenging scenarios identified from Exp 2, and Exp 7 assigns a distinct expert to each of the 44 scenarios in Bench2Drive. We observe that excessively increasing the number of experts (Exp 5-6) negatively affects performance due to the induced load imbalance among experts. Thus, an appropriate balance in the number of specialized experts is crucial for optimal driving performance.

\begin{table}[h!]
\vspace{-5pt}
\small
\centering
\caption{\textbf{Router Accuracy}. The vision router and action router accuracy in Bench2Drive-Base validation set.\label{tab:router_acc}}
    \begin{tabular}{c|c}
    \toprule
    \textbf{Router} & \textbf{Accuracy$(\%)\uparrow$} \\ \hline
    Vision Router & 88.85  \\
    Action Router & 65.40  \\ \bottomrule
    \end{tabular}
\vspace{-5pt}
\end{table}

\begin{table}[h!]
\vspace{-5pt}
\caption{\textbf{Token vs Trajectory Level Action MoE.}}
\label{tab:style_action}
\resizebox{0.98\linewidth}{!}{
\begin{tabular}{c|c|c|c|c}
\toprule
\textbf{Style} & \textbf{Share} & \textbf{Non-Share} & \textbf{DS$\uparrow$} & \textbf{SR$(\%)\uparrow$} \\ \hline
Token-Level &   3    &      6     &  65.62  &  32.27	  \\
\hline
Traj-Level &   3    &      6     &  73.88	& 48.64  \\ 
\bottomrule
\end{tabular}}
\vspace{-6pt}
\end{table}

\begin{table}[h!]
    \vspace{-5pt}
    \centering
    \caption{\textbf{Ablation Study in Action MoE}. Compare various configurations of non-share expert numbers within Action MoE. \label{tab:ablation_action_moe}}
    \centering
    \resizebox{0.40\textwidth}{!}{
    \begin{tabular}{c|c|c|c|c|c}
    \toprule
    \textbf{Exp} & \textbf{Share} & \textbf{Non-share} & \textbf{Supervision} & \textbf{DS$\uparrow$} & \textbf{SR$(\%)\uparrow$} \\ \hline
    1&1&5 &$\checkmark$& 73.81 & 47.73 \\
    \rowcolor{gray!20} 2&1&6 &$\checkmark$& 74.22 & 48.64  \\
    3&1&6 &$\times$& 70.38 & 45.00 \\
    4&3&6 &$\checkmark$& 73.88 & 48.64  \\
    5&5&6 &$\checkmark$& 73.46 & 47.73  \\
    6&1&13 &$\checkmark$& 70.88 & 44.50 \\
    7&1&44 &$\checkmark$& 68.22 & 43.18 \\
    \bottomrule
    \end{tabular}
    }
\vspace{-6pt}
\end{table}

\noindent
\textbf{Drive-$\boldsymbol{\pi}_{\boldsymbol{0}}$ vs DriveMoE.}
We conduct ablation studies to evaluate the individual contributions of the Vision MoE and Action MoE within our DriveMoE framework. As shown in Table~\ref{pi0moe}, removing either the Vision MoE or the Action MoE leads to a noticeable decline in both driving score and success rate. Compared to the Drive-$\boldsymbol{\pi}_{\boldsymbol{0}}$, DriveMoE substantially improves driving performance, highlighting the complementary effectiveness of both MoE modules.

\begin{table}[h!]
\vspace{-5pt}
    \centering
    \centering
    \caption{\textbf{Drive-$\boldsymbol{\pi}_{\boldsymbol{0}}$ vs DriveMoE}. Evaluate the Vision MoE and Action MoE. ''w/o'' denotes removing the respective modules.\label{pi0moe}}
    \resizebox{0.3\textwidth}{!}{
    \begin{tabular}{l|c|c}
    \toprule
    \textbf{Method}        & \textbf{DS$\uparrow$} & \textbf{SR$(\%)\uparrow$} \\ \hline
    Drive-$\boldsymbol{\pi}_{\boldsymbol{0}}$ & 55.85 & 30.00 \\
    w/o Vision MoE & 68.68 & 42.45 \\
    w/o Action MoE & 67.31 & 40.56 \\ 
    DriveMoE & 74.22 & 48.64 \\ \bottomrule
    \end{tabular}}
\vspace{-10pt}
\end{table}

\noindent

\vspace{-6pt}
\section{Conclusion}
We propose DriveMoE, a novel end-to-end autonomous driving framework built upon Drive-$\boldsymbol{\pi}_{\boldsymbol{0}}$, which integrates Mixture-of-Experts(MoE) into both vision and action components. DriveMoE effectively addresses challenges inherent in existing VLA models by dynamically selecting relevant camera views through a Scene-Specialized Vision MoE, and by employing a Skill-Specialized Action MoE that activates expert modules tailored to specific driving behaviors. Extensive evaluations on the Bench2Drive benchmark show that DriveMoE achieves state-of-the-art performance, significantly enhancing computational efficiency and robustness to rare, safety-critical driving scenarios.

\clearpage
{
    \small
    \bibliographystyle{ieeenat_fullname}
    \bibliography{main}
}

\input{X_suppl}

\end{document}

%% file: X_suppl.tex
\clearpage
\setcounter{page}{1}
\maketitlesupplementary

\section{Related Work}
\label{sec:related_work}
\subsection{VLM/VLA in End-to-end Autonomous Driving}

The advancement of Large Language Models (LLMs) has significantly accelerated the development of Vision-Language Models (VLMs) for autonomous driving. Leveraging powerful generalization, open-set reasoning, and scalability, these models have become influential paradigms for end-to-end driving tasks. 

DriveGPT-4~\cite{xu2024drivegpt4} pioneers the integration of multimodal LLMs into end-to-end driving by introducing a vision–language–action framework that jointly performs control prediction and natural-language explanation. DriveLM~\cite{sima2023drivelm} further explores the reasoning aspect of driving intelligence through graph visual question answering (GVQA). LMDrive~\cite{shao2023lmdrive} formulates perception and planning tasks as sequences of discrete tokens, enabling better interpretability and facilitating cross-domain knowledge transfer. LeapAD~\cite{mei2024continuously} adopts a data-driven and knowledge-driven Vision-Language Model (VLM) approach. SimLingo~\cite{renz2025simlingo} proposes a fully closed-loop VLA architecture aligning language reasoning with low-level control. To enhance learning efficiency, it introduces a data-bucket sampling scheme that focuses on diverse and high-risk scenarios, avoiding overfitting to trivial straight-driving cases. 

However, Existing approaches predominantly rely on discrete token-based for driving policy learning, without exploring continuous tokenization or diffusion-based policies that could bridge vision-language understanding and continuous control. Moreover, despite the increasing interest in multimodal skill modeling for complex driving scenes, no prior work has investigated Mixture-of-Experts (MoE) architectures to enhance specialization and generalization across diverse driving domains.

To address this limitation, the embodied AI community has proposed vision-language-action (VLA) models that represent actions as continuous variables instead of discrete tokens. Methods such as OpenVLA~\cite{kim2024openvla}, Diffusion Policy~\cite{chi2023diffusion} and $\pi_0$~\cite{black2024pi_0} demonstrate strong performance by modeling continuous action distributions through sequence prediction and global optimization. Nevertheless, these approaches often rely on task-specific policies or instruction-conditioned models, which struggle to generalize across the long-tail distribution of behaviors seen in complex driving environments.

\subsection{Mixture-of-Experts in Large Language Models}
Recent progress in Mixture-of-Experts (MoE) architectures has demonstrated remarkable efficiency and scalability in large language models (LLMs). Sparse Mixture-of-Experts (MoE) architectures have become a mainstream approach for scaling LLMs. By replacing the standard feedforward layers in Transformers with expert modules, models like DeepSeekMoE~\cite{dai2024deepseekmoe} and Mixtral-8x7B~\cite{dai2024deepseekmoe} improve task specialization and representation capacity while maintaining inference efficiency through conditional computation. In robotics, MoE architectures have also been used to address task heterogeneity and long-tailed data distributions. For example, MENTOR~\cite{huang2024mentor} replaces the MLP backbone with MoE layers to enable gradient routing among modular experts, helping mitigate gradient interference in multi-task learning. Despite promising results in language modeling and robot policy learning, the use of MoE in end-to-end autonomous driving remains underexplored.  

Motivated by these gaps, our metheds is the first end-to-end autonomous driving framework that integrates MoE at 
both the vision and action levels.

\section{Conditional Flow Matching Loss}
\label{sec:flow_loss}
Following prior work~\cite{black2024pi_0, lipman2022flow, liu2022rectified}, our method predicts future action trajectories in a denoising manner using a conditional flow matching loss,

\begin{equation}
\small
L^{\tau}(\theta) = \mathbb{E}_{p(\mathbf{A}_t \mid \mathbf{o}_t),\, q(\mathbf{A}_t^{\tau} \mid \mathbf{A}_t)} 
\left\| 
\mathbf{v}_{\theta}(\mathbf{A}_t^{\tau}, \mathbf{o}_t) - \mathbf{u}(\mathbf{A}_t^{\tau} \mid \mathbf{A}_t) 
\right\|^2
\end{equation}

where subscripts denote timesteps and superscripts denote flow matching timesteps, with $\tau \in [0,1]$. We sample noisy actions $\mathbf{A}_t^\tau = \tau \mathbf{A}t + (1 - \tau)\epsilon$ and train the network to output a denoising flow $\mathbf{v}_\theta(\mathbf{A}_t^\tau, o_t)$ that matches the ground-truth direction $\mathbf{u}(\mathbf{A}_t^{\tau} \mid \mathbf{A}_t) = \epsilon - \mathbf{A}_t$. This formulation enables the model to learn the underlying trajectory distribution by aligning with a continuous stochastic process, rather than relying on pointwise supervision. It is particularly suitable for multimodal or uncertain planning scenarios in autonomous driving, where modeling smooth action trajectories is critical.

\section{Implementation Details}
\textbf{Vision Routing Annotations:} We introduce additional camera-view importance annotations into the Bench2Drive~\cite{jia2024bench} dataset. This annotation approach is both inexpensive and straightforward, yet it significantly improves model performance through efficient and effective utilization of multi-camera inputs. The details about camera annotation rules refer to Appendix~\ref{sup:moe_label}.

\noindent
\textbf{Action Routing Annotations:} We maintain skill definitions consistent with   Bench2Drive~\cite{jia2024bench} setup. There are five driving skills: Merging, Overtaking, Emergency Brake, Give Way, and Traffic Sign.

\noindent
\textbf{Drive-$\boldsymbol{\pi}_{\boldsymbol{0}}$:} We utilize 2 sequential front-view images as input to our model to effectively estimate the velocities of surrounding traffic agents. Additionally, the input state incorporates both current and historical information, including position, velocity, acceleration, and heading angle, enabling the model to predict 10 future waypoints accurately.

\noindent
\textbf{DriveMoE:} We utilize 2 sequential front-view images combined with a dynamically selected camera view as inputs to our model. The sequential front-view images primarily capture temporal changes to model the velocities of surrounding traffic agents, while the dynamic view is obtained by selecting the Top-1 view from the vision router, which enhances spatial perception according to driving context. The input state representation remains consistent with the $\pi_0$ framework, including current and historical position, velocity, acceleration, and heading angle information. In the action model, we employ 1 shared expert and 6 non-shared experts. During the training and inference, the top-3 experts selected by the action router are utilized to generate the final trajectory prediction consisting of 10 future waypoints. We adopt a two-stage post-training strategy for our model:

\noindent
\textbf{Training Stage 1}.  We train the model for 12 epochs. The Vision-Language Model (VLM) component is initialized from the pretrained weights of Paligemma-3b-pt-224~\cite{beyer2024paligemma}. The VLM and Action MoE experts are optimized separately using two optimizers, both configured as follows: learning rate = $5\times10^{-5}$, and warmup steps enabled. Gradient clipping is applied with a maximum gradient norm of 1.0. Gradient accumulation is used to simulate a batch size of 128. To balance different loss components effectively, we set the vision router loss weight $\lambda_0$ to 10.0, action router loss weight $\lambda_2$ to 10.0, flow matching loss weight $\lambda_1$ to 1.0. 

\noindent
\textbf{Training Stage 2}. We continue training for an additional 6 epochs, initializing from the checkpoint obtained at the end of Stage 1. In this stage, input camera views and action experts are dynamically selected based on outputs from the routers.the vision router loss weight $\lambda_0$ to 5.0, action router loss weight $\lambda_2$ to 5.0, flow matching loss weight $\lambda_1$ to 1.0, emphasizing trajectory learning. Other hyperparameters remain consistent with Stage 1.

\noindent
\textbf{PID Controller}.  All methods use the same PID controller for fair comparison in closed-loop evaluation. The PID controller module takes as input the current vehicle speed and the future trajectory predicted by the model, consisting of 10 waypoints, and outputs throttle, brake, and steering angle commands. Specifically, for the steering control, the PID gains are: $K_{P}^{\text{turn}}$ = 1.25, $K_{I}^{\text{turn}}$ = 0.75, $K_{D}^{\text{turn}}$ = 0.3
For speed control, the PID gains are: $K_{P}^{\text{speed}}$ = 5.0, $K_{I}^{\text{speed}}$ = 0.5, $K_{D}^{\text{speed}}$ = 1.0. The desired vehicle speed is computed from the 7th waypoint of the predicted trajectory, whereas the steering angle is determined using the 10th waypoint. This configuration ensures stable and responsive vehicle control aligned with the model’s trajectory predictions.

\section{Discussion on Unsupervised View Selection and Knowledge Distillation}

About unsupervised view selection, unsupervised camera view learning could further reduce annotation costs and improve generalization. Our supervised annotations are inexpensive (simple heuristics from trajectory/map) and provide stable training for initial Vision MoE exploration. Nevertheless, the methods DySS~\cite{yasarla2025dyss} and LightVLA~\cite{jiang2025better} offer valuable insights that could inspire image-level pruning. Potential extensions include image-level token pruning after view selection.

About Knowledge Distillation, Combining DriveMoE with knowledge distillation~\cite{hegde2025distilling} is a promising direction—our architecture could serve as a teacher model to distill skill-specific compact students for deployment.

\section{Annotation for Router}
\label{sup:moe_label}

\textbf{Vision Router:} We developed a set of heuristic rules based on annotation information from the Bench2Drive dataset to identify special driving scenarios, enabling effective camera-view-level supervision. We select camera views contextually, defaulting to the rear view if no critical view is identified. The Camera Annotation Rules are, 
\begin{itemize}[leftmargin=*]
    \item \textbf{Intersection Turning:} When the ego-vehicle is required to turn at an intersection (i.e., \texttt{is\_in\_junction} is true and the current command is either ``turn left'' or ``turn right''), we annotate the front-side camera view pointing toward the intended exit of the intersection.

    \item \textbf{Lane Change:} When a lane change is required, identified by conditions such as the current command being ``change left'' or ``change right,'' an obstacle appearing within a certain distance ahead in the current lane, or the ego-vehicle not being in the target lane, the annotation depends on lane direction:
    \begin{itemize}
        \item If the target lane is in the same direction as the ego-vehicle's current movement, we annotate the corresponding rear-side camera.
        \item If the ego-vehicle must temporarily occupy the opposing lane, we annotate the corresponding front-side camera.
    \end{itemize}

    \item \textbf{Highway Merging and Cut-in:} In scenarios such as highway merging or vehicle cut-ins (scenario labeled as ``merging'' or ``cut-in''), we determine the merging location based on the ego-vehicle's lane position and distance to the junction, annotating the side camera facing the merging location.

    \item \textbf{Yielding to Emergency Vehicles:} If a high-speed emergency vehicle is present in the scenario, the ego-vehicle must yield, and we annotate the camera facing the direction of the approaching emergency vehicle.
\end{itemize}

\noindent\textbf{Action Router:} As shown in Table~\ref{tab:skill}, Bench2Drive~\cite{jia2024bench} divides 44 scenarios into 5 skills.
\begin{table*}[t]
\caption{\textbf{Skill Set \& Scenarios}\label{tab:skill}}
\begin{tabular}{l|p{13cm}}
\toprule
\textbf{Skill}  & \textbf{Scenario}                                                                                        \\ \midrule
Merging         & CrossingBicycleFlow, EnterActorFlow, HighwayExit, InterurbanActorFlow, HighwayCutIn, InterurbanAdvancedActorFlow, MergerIntoSlowTrafficV2, MergeIntoSlowTraffic, NonSignalizedJunctionLeftTurn, NonSignalizedJunctionRightTurn, NonSignalizedJunctionLeftTurnEnterFlow, ParkingExit, LaneChange, SignalizedJunctionLeftTurn, SignalizedJunctionRightTurn, SignalizedJunctionLeftTurnEnterFlow                                                                   \\ \midrule
Overtaking      & Accident, AccidentTwoWays, ConstructionObstacle, ConstructionObstacleTwoWays, HazardAtSideLaneTwoWays, HazardAtSideLane, ParkedObstacleTwoWays, ParkedObstacle, VehicleOpenDoorTwoWays                                                                \\ \midrule
Emergency Brake & BlockedIntersection, DynamicObjectCrossing, HardBreakRoute, OppositeVehicleTakingPriority, OppositeVehicleRunningRedLight, ParkingCutIn, PedestrianCrossing, ParkingCrossingPedestrian, StaticCutIn, VehicleTurningRoute, VehicleTurningRoutePedestrian, ControlLoss                                         \\ \midrule
Give Way        & InvadingTurn, YieldToEmergencyVehicle                                                                     \\ \midrule
Traffic Sign    & EnterActorFlow, CrossingBicycleFlow, NonSignalizedJunctionLeftTurn, NonSignalizedJunctionRightTurn, NonSignalizedJunctionLeftTurnEnterFlow, OppositeVehicleTakingPriority, OppositeVehicleRunningRedLight, PedestrianCrossing, SignalizedJunctionLeftTurn, SignalizedJunctionRightTurn, SignalizedJunctionLeftTurnEnterFlow, TJunction, VanillaNonSignalizedTurn, VanillaSignalizedTurnEncounterGreenLight, VanillaSignalizedTurnEncounterRedLight, VanillaNonSignalizedTurnEncounterStopsign, VehicleTurningRoute, VehicleTurningRoutePedestrian \\ \bottomrule
\end{tabular}
\end{table*}

\section{Experiment on nuScenes}
\begin{table}[H]
\centering
\caption{Open-loop planning performance in nuScenes}
\label{tab:nus}
\resizebox{0.48\textwidth}{!}{
\begin{tabular}{l|cccc|cccc}
\toprule
\multirow{2}{*}{Method}
& \multicolumn{4}{c|}{L2 (m)$\downarrow$}
& \multicolumn{4}{c}{Collision (\%)$\downarrow$} \\
\cline{2-9}
& 1s & 2s & 3s & Avg.
& 1s & 2s & 3s & Avg. \\
\midrule
UniAD~\cite{hu2023planning}
& 0.48 & 0.74 & 1.07 & \cellcolor{gray!20}{0.76}
& 0.12 & \textbf{0.13} & 0.28 & \cellcolor{gray!20}{\textbf{0.17}} \\
VAD-Base~\cite{jiang2023vad}
& \textbf{0.41} & \textbf{0.70} & \textbf{1.05} & \cellcolor{gray!20}{\textbf{0.72}}
& \textbf{0.07} & 0.17 & 0.41 & \cellcolor{gray!20}{0.22} \\
Drive-$\boldsymbol{\pi}_{\boldsymbol{0}}$
& 0.51 & 0.73 & 1.11 & \cellcolor{gray!20}{0.78}
& 0.14 & 0.18 & 0.39 & \cellcolor{gray!20}{0.24} \\
DriveMoE
& 0.45 & \textbf{0.70} & 1.08 & \cellcolor{gray!20}{0.74}
& 0.11 & 0.15 & \textbf{0.26} & \cellcolor{gray!20}{\textbf{0.17}} \\
\bottomrule
\end{tabular}}
\end{table}

Following the same skill definitions as Bench2Drive, DriveMoE achieves competitive L2 error while significantly reducing collision rate (Table~\ref{tab:nus}), demonstrating strong generalization to real-world scenarios. Same to~\cite{fu2025orion}, open-loop evaluation cannot reflect driving performance.

\section{Hyperparameters \& Efficienc}
As shown in Table~\ref{tab:cost}, the increased cost primarily stems from processing one additional camera view and the MoE modules with Top-3 expert activation. We will clarify this in our revision.

\begin{table}[H]
\centering
\caption{Comparison of model scale and inference cost.}
\label{tab:cost}
\resizebox{0.50\textwidth}{!}{
\begin{tabular}{l|cccccccc}
\toprule
Method & View Num & View & DS$\uparrow$ & SR(\%)$\uparrow$ & Parameters (M) & FLOPs (G) & Latency(ms)$\downarrow$ & Training Cost \\
\midrule
Drive-$\boldsymbol{\pi}_{\boldsymbol{0}}$ & 2 & Fixed & 63.26 & 31.82 & 2606 & 3400 & 240 & 80 GPU-hours \\
Drive-$\boldsymbol{\pi}_{\boldsymbol{0}}$ & 6 & Fixed & 62.27 & 31.36 & 2606 & 7576 & 700 & 320 GPU-hours \\
DriveMoE & 2 & Dynamic & \textbf{74.22} & \textbf{48.64} & 3008 & 3896 & 260 & 120/80 GPU-hours \\
\bottomrule
\end{tabular}
}
\end{table}

%% file: main.bib
@String(CVPR= {IEEE Conf. Comput. Vis. Pattern Recog.})

@String(ICCV= {Int. Conf. Comput. Vis.})

@String(CVPR  = {CVPR})

@String(ICCV  = {ICCV})

@article{Chitta2023PAMI,
  author = {Chitta, Kashyap and
            Prakash, Aditya and
            Jaeger, Bernhard and
            Yu, Zehao and
            Renz, Katrin and
            Geiger, Andreas},
  title = {TransFuser: Imitation with Transformer-Based Sensor Fusion for Autonomous Driving},
  journal = {TPAMI},
  year = {2023},
}

@inproceedings{hu2023planning,
  title={Planning-oriented autonomous driving},
  author={Hu, Yihan and Yang, Jiazhi and Chen, Li and Li, Keyu and Sima, Chonghao and Zhu, Xizhou and Chai, Siqi and Du, Senyao and Lin, Tianwei and Wang, Wenhai and others},
  booktitle={CVPR},
  pages={17853--17862},
  year={2023}
}

@article{jiang2023vad,
  title={VAD: Vectorized Scene Representation for Efficient Autonomous Driving},
  author={Jiang, Bo and Chen, Shaoyu and Xu, Qing and Liao, Bencheng and Chen, Jiajie and Zhou, Helong and Zhang, Qian and Liu, Wenyu and Huang, Chang and Wang, Xinggang},
  journal={ICCV},
  year={2023}
}

@misc{shao2023lmdrive,
      title={LMDrive: Closed-Loop End-to-End Driving with Large Language Models}, 
      author={Hao Shao and Yuxuan Hu and Letian Wang and Steven L. Waslander and Yu Liu and Hongsheng Li},
      year={2023},
      eprint={2312.07488},
      archivePrefix={arXiv},
      primaryClass={cs.CV}
}

@article{sima2023drivelm,
  title={DriveLM: Driving with Graph Visual Question Answering},
  author={Sima, Chonghao and Renz, Katrin and Chitta, Kashyap and Chen, Li and Zhang, Hanxue and Xie, Chengen and Luo, Ping and Geiger, Andreas and Li, Hongyang},
  journal={arXiv preprint arXiv:2312.14150},
  year={2023}
}

@misc{renz2024carllavavisionlanguagemodels,
      title={CarLLaVA: Vision language models for camera-only closed-loop driving}, 
      author={Katrin Renz and Long Chen and Ana-Maria Marcu and Jan Hünermann and Benoit Hanotte and Alice Karnsund and Jamie Shotton and Elahe Arani and Oleg Sinavski},
      year={2024},
      eprint={2406.10165},
      archivePrefix={arXiv},
      primaryClass={cs.CV},
      url={https://arxiv.org/abs/2406.10165}, 
}

@article{cai2025survey,
  title={A Survey on Mixture of Experts in Large Language Models},
  author={Cai, Weilin and Jiang, Juyong and Wang, Fan and Tang, Jing and Kim, Sunghun and Huang, Jiayi},
  journal={IEEE Transactions on Knowledge and Data Engineering},
  year={2025},
  publisher={IEEE}
}

@article{wan2023efficient,
  title={Efficient large language models: A survey},
  author={Wan, Zhongwei and Wang, Xin and Liu, Che and Alam, Samiul and Zheng, Yu and others},
  journal={arXiv preprint arXiv:2312.03863},
  volume={1},
  year={2023},
  publisher={no}
}

@inproceedings{zhu2024llama,
  title={Llama-moe: Building mixture-of-experts from llama with continual pre-training},
  author={Zhu, Tong and Qu, Xiaoye and Dong, Daize and Ruan, Jiacheng and Tong, Jingqi and He, Conghui and Cheng, Yu},
  booktitle={Proceedings of the 2024 Conference on Empirical Methods in Natural Language Processing},
  pages={15913--15923},
  year={2024}
}

@article{bai2023qwen,
  title={Qwen technical report},
  author={Bai, Jinze and Bai, Shuai and Chu, Yunfei and Cui, Zeyu and Dang, Kai and Deng, Xiaodong and Fan, Yang and Ge, Wenbin and Han, Yu and Huang, Fei and others},
  journal={arXiv preprint arXiv:2309.16609},
  year={2023}
}

@misc{openai2024gpt4technicalreport,
      title={GPT-4 Technical Report}, 
      author={OpenAI and Josh Achiam and Steven Adler and Sandhini Agarwal and Lama Ahmad and Ilge Akkaya and Florencia Leoni Aleman and Diogo Almeida and Janko Altenschmidt and Sam Altman and Shyamal Anadkat and Red Avila and Igor Babuschkin and Suchir Balaji and Valerie Balcom and Paul Baltescu and Haiming Bao and Mohammad Bavarian and Jeff Belgum and Irwan Bello and Jake Berdine and Gabriel Bernadett-Shapiro and Christopher Berner and Lenny Bogdonoff and Oleg Boiko and Madelaine Boyd and Anna-Luisa Brakman and Greg Brockman and Tim Brooks and Miles Brundage and Kevin Button and Trevor Cai and Rosie Campbell and Andrew Cann and Brittany Carey and Chelsea Carlson and Rory Carmichael and Brooke Chan and Che Chang and Fotis Chantzis and Derek Chen and Sully Chen and Ruby Chen and Jason Chen and Mark Chen and Ben Chess and Chester Cho and Casey Chu and Hyung Won Chung and Dave Cummings and Jeremiah Currier and Yunxing Dai and Cory Decareaux and Thomas Degry and Noah Deutsch and Damien Deville and Arka Dhar and David Dohan and Steve Dowling and Sheila Dunning and Adrien Ecoffet and Atty Eleti and Tyna Eloundou and David Farhi and Liam Fedus and Niko Felix and Simón Posada Fishman and Juston Forte and Isabella Fulford and Leo Gao and Elie Georges and Christian Gibson and Vik Goel and Tarun Gogineni and Gabriel Goh and Rapha Gontijo-Lopes and Jonathan Gordon and Morgan Grafstein and Scott Gray and Ryan Greene and Joshua Gross and Shixiang Shane Gu and Yufei Guo and Chris Hallacy and Jesse Han and Jeff Harris and Yuchen He and Mike Heaton and Johannes Heidecke and Chris Hesse and Alan Hickey and Wade Hickey and Peter Hoeschele and Brandon Houghton and Kenny Hsu and Shengli Hu and Xin Hu and Joost Huizinga and Shantanu Jain and Shawn Jain and Joanne Jang and Angela Jiang and Roger Jiang and Haozhun Jin and Denny Jin and Shino Jomoto and Billie Jonn and Heewoo Jun and Tomer Kaftan and Łukasz Kaiser and Ali Kamali and Ingmar Kanitscheider and Nitish Shirish Keskar and Tabarak Khan and Logan Kilpatrick and Jong Wook Kim and Christina Kim and Yongjik Kim and Jan Hendrik Kirchner and Jamie Kiros and Matt Knight and Daniel Kokotajlo and Łukasz Kondraciuk and Andrew Kondrich and Aris Konstantinidis and Kyle Kosic and Gretchen Krueger and Vishal Kuo and Michael Lampe and Ikai Lan and Teddy Lee and Jan Leike and Jade Leung and Daniel Levy and Chak Ming Li and Rachel Lim and Molly Lin and Stephanie Lin and Mateusz Litwin and Theresa Lopez and Ryan Lowe and Patricia Lue and Anna Makanju and Kim Malfacini and Sam Manning and Todor Markov and Yaniv Markovski and Bianca Martin and Katie Mayer and Andrew Mayne and Bob McGrew and Scott Mayer McKinney and Christine McLeavey and Paul McMillan and Jake McNeil and David Medina and Aalok Mehta and Jacob Menick and Luke Metz and Andrey Mishchenko and Pamela Mishkin and Vinnie Monaco and Evan Morikawa and Daniel Mossing and Tong Mu and Mira Murati and Oleg Murk and David Mély and Ashvin Nair and Reiichiro Nakano and Rajeev Nayak and Arvind Neelakantan and Richard Ngo and Hyeonwoo Noh and Long Ouyang and Cullen O'Keefe and Jakub Pachocki and Alex Paino and Joe Palermo and Ashley Pantuliano and Giambattista Parascandolo and Joel Parish and Emy Parparita and Alex Passos and Mikhail Pavlov and Andrew Peng and Adam Perelman and Filipe de Avila Belbute Peres and Michael Petrov and Henrique Ponde de Oliveira Pinto and Michael and Pokorny and Michelle Pokrass and Vitchyr H. Pong and Tolly Powell and Alethea Power and Boris Power and Elizabeth Proehl and Raul Puri and Alec Radford and Jack Rae and Aditya Ramesh and Cameron Raymond and Francis Real and Kendra Rimbach and Carl Ross and Bob Rotsted and Henri Roussez and Nick Ryder and Mario Saltarelli and Ted Sanders and Shibani Santurkar and Girish Sastry and Heather Schmidt and David Schnurr and John Schulman and Daniel Selsam and Kyla Sheppard and Toki Sherbakov and Jessica Shieh and Sarah Shoker and Pranav Shyam and Szymon Sidor and Eric Sigler and Maddie Simens and Jordan Sitkin and Katarina Slama and Ian Sohl and Benjamin Sokolowsky and Yang Song and Natalie Staudacher and Felipe Petroski Such and Natalie Summers and Ilya Sutskever and Jie Tang and Nikolas Tezak and Madeleine B. Thompson and Phil Tillet and Amin Tootoonchian and Elizabeth Tseng and Preston Tuggle and Nick Turley and Jerry Tworek and Juan Felipe Cerón Uribe and Andrea Vallone and Arun Vijayvergiya and Chelsea Voss and Carroll Wainwright and Justin Jay Wang and Alvin Wang and Ben Wang and Jonathan Ward and Jason Wei and CJ Weinmann and Akila Welihinda and Peter Welinder and Jiayi Weng and Lilian Weng and Matt Wiethoff and Dave Willner and Clemens Winter and Samuel Wolrich and Hannah Wong and Lauren Workman and Sherwin Wu and Jeff Wu and Michael Wu and Kai Xiao and Tao Xu and Sarah Yoo and Kevin Yu and Qiming Yuan and Wojciech Zaremba and Rowan Zellers and Chong Zhang and Marvin Zhang and Shengjia Zhao and Tianhao Zheng and Juntang Zhuang and William Zhuk and Barret Zoph},
      year={2024},
      eprint={2303.08774},
      archivePrefix={arXiv},
      primaryClass={cs.CL},
      url={https://arxiv.org/abs/2303.08774}, 
}

@misc{chen2024asynchronouslargelanguagemodel,
      title={Asynchronous Large Language Model Enhanced Planner for Autonomous Driving}, 
      author={Yuan Chen and Zi-han Ding and Ziqin Wang and Yan Wang and Lijun Zhang and Si Liu},
      year={2024},
      eprint={2406.14556},
      archivePrefix={arXiv},
      primaryClass={cs.RO},
      url={https://arxiv.org/abs/2406.14556}, 
}

@inproceedings{qi2024gpt4point,
  title={Gpt4point: A unified framework for point-language understanding and generation},
  author={Qi, Zhangyang and Fang, Ye and Sun, Zeyi and Wu, Xiaoyang and Wu, Tong and Wang, Jiaqi and Lin, Dahua and Zhao, Hengshuang},
  booktitle={Proceedings of the IEEE/CVF Conference on Computer Vision and Pattern Recognition},
  pages={26417--26427},
  year={2024}
}

@inproceedings{li2023blip,
  title={Blip-2: Bootstrapping language-image pre-training with frozen image encoders and large language models},
  author={Li, Junnan and Li, Dongxu and Savarese, Silvio and Hoi, Steven},
  booktitle={International conference on machine learning},
  pages={19730--19742},
  year={2023},
  organization={PMLR}
}

@article{black2024pi_0,
  title={$pi\_0$: A Vision-Language-Action Flow Model for General Robot Control},
  author={Black, Kevin and Brown, Noah and Driess, Danny and Esmail, Adnan and Equi, Michael and Finn, Chelsea and Fusai, Niccolo and Groom, Lachy and Hausman, Karol and Ichter, Brian and others},
  journal={arXiv preprint arXiv:2410.24164},
  year={2024}
}

@article{intelligence2025pi_,
  title={$pi\_0.5$: a Vision-Language-Action Model with Open-World Generalization},
  author={Intelligence, Physical and Black, Kevin and Brown, Noah and Darpinian, James and Dhabalia, Karan and Driess, Danny and Esmail, Adnan and Equi, Michael and Finn, Chelsea and Fusai, Niccolo and others},
  journal={arXiv preprint arXiv:2504.16054},
  year={2025}
}

@article{lipman2022flow,
  title={Flow matching for generative modeling},
  author={Lipman, Yaron and Chen, Ricky TQ and Ben-Hamu, Heli and Nickel, Maximilian and Le, Matt},
  journal={arXiv preprint arXiv:2210.02747},
  year={2022}
}

@article{xu2024drivegpt4,
  title={Drivegpt4: Interpretable end-to-end autonomous driving via large language model},
  author={Xu, Zhenhua and Zhang, Yujia and Xie, Enze and Zhao, Zhen and Guo, Yong and Wong, Kwan-Yee K and Li, Zhenguo and Zhao, Hengshuang},
  journal={IEEE Robotics and Automation Letters},
  year={2024},
  publisher={IEEE}
}

@article{chi2023diffusion,
  title={Diffusion policy: Visuomotor policy learning via action diffusion},
  author={Chi, Cheng and Xu, Zhenjia and Feng, Siyuan and Cousineau, Eric and Du, Yilun and Burchfiel, Benjamin and Tedrake, Russ and Song, Shuran},
  journal={The International Journal of Robotics Research},
  pages={02783649241273668},
  year={2023},
  publisher={SAGE Publications Sage UK: London, England}
}

@article{kim2024openvla,
  title={Openvla: An open-source vision-language-action model},
  author={Kim, Moo Jin and Pertsch, Karl and Karamcheti, Siddharth and Xiao, Ted and Balakrishna, Ashwin and Nair, Suraj and Rafailov, Rafael and Foster, Ethan and Lam, Grace and Sanketi, Pannag and others},
  journal={arXiv preprint arXiv:2406.09246},
  year={2024}
}

@article{dai2024deepseekmoe,
  title={Deepseekmoe: Towards ultimate expert specialization in mixture-of-experts language models},
  author={Dai, Damai and Deng, Chengqi and Zhao, Chenggang and Xu, RX and Gao, Huazuo and Chen, Deli and Li, Jiashi and Zeng, Wangding and Yu, Xingkai and Wu, Yu and others},
  journal={arXiv preprint arXiv:2401.06066},
  year={2024}
}

@article{jiang2024mixtral,
  title={Mixtral of experts},
  author={Jiang, Albert Q and Sablayrolles, Alexandre and Roux, Antoine and Mensch, Arthur and Savary, Blanche and Bamford, Chris and Chaplot, Devendra Singh and Casas, Diego de las and Hanna, Emma Bou and Bressand, Florian and others},
  journal={arXiv preprint arXiv:2401.04088},
  year={2024}
}

@article{huang2024mentor,
  title={MENTOR: Mixture-of-Experts Network with Task-Oriented Perturbation for Visual Reinforcement Learning},
  author={Huang, Suning and Zhang, Zheyu and Liang, Tianhai and Xu, Yihan and Kou, Zhehao and Lu, Chenhao and Xu, Guowei and Xue, Zhengrong and Xu, Huazhe},
  journal={arXiv preprint arXiv:2410.14972},
  year={2024}
}

@article{zheng2024genad,
    title={GenAD: Generative End-to-End Autonomous Driving},
    author={Zheng, Wenzhao and Song, Ruiqi and Guo, Xianda and Zhang, Chenming and Chen, Long},
    journal={arXiv preprint arXiv: 2402.11502},
    year={2024}
}

@article{song2025don,
  title={Don't Shake the Wheel: Momentum-Aware Planning in End-to-End Autonomous Driving},
  author={Song, Ziying and Jia, Caiyan and Liu, Lin and Pan, Hongyu and Zhang, Yongchang and Wang, Junming and Zhang, Xingyu and Xu, Shaoqing and Yang, Lei and Luo, Yadan},
  journal={arXiv preprint arXiv:2503.03125},
  year={2025}
}

@article{wang2025diffad,
  title={DiffAD: A Unified Diffusion Modeling Approach for Autonomous Driving},
  author={Wang, Tao and Zhang, Cong and Qu, Xingguang and Li, Kun and Liu, Weiwei and Huang, Chang},
  journal={arXiv preprint arXiv:2503.12170},
  year={2025}
}

@article{li2025end,
  title={End-to-End Driving with Online Trajectory Evaluation via BEV World Model},
  author={Li, Yingyan and Wang, Yuqi and Liu, Yang and He, Jiawei and Fan, Lue and Zhang, Zhaoxiang},
  journal={arXiv preprint arXiv:2504.01941},
  year={2025}
}

@article{beyer2024paligemma,
  title={Paligemma: A versatile 3b vlm for transfer},
  author={Beyer, Lucas and Steiner, Andreas and Pinto, Andr{\'e} Susano and Kolesnikov, Alexander and Wang, Xiao and Salz, Daniel and Neumann, Maxim and Alabdulmohsin, Ibrahim and Tschannen, Michael and Bugliarello, Emanuele and others},
  journal={arXiv preprint arXiv:2407.07726},
  year={2024}
}

@inproceedings{wu2022trajectoryguided,
 title={Trajectory-guided Control Prediction for End-to-end Autonomous Driving: A Simple yet Strong Baseline}, 
 author={Penghao Wu and Xiaosong Jia and Li Chen and Junchi Yan and Hongyang Li and Yu Qiao},
 booktitle={NeurIPS},
 year={2022},
}

@inproceedings{jia2023thinktwice,
  title={Think Twice before Driving: Towards Scalable Decoders for End-to-End Autonomous Driving},
  author={Jia, Xiaosong and Wu, Penghao and Chen, Li and Xie, Jiangwei and He, Conghui and Yan, Junchi and Li, Hongyang},
  booktitle={CVPR},
  year={2023}
}

@inproceedings{jia2023driveadapter,
  title={DriveAdapter: Breaking the Coupling Barrier of Perception and Planning in End-to-End Autonomous Driving},
  author={Jia, Xiaosong and Gao, Yulu and Chen, Li and Yan, Junchi and Liu, Patrick Langechuan and Li, Hongyang},
  booktitle={ICCV},
  year={2023}
}

@InProceedings{Jaeger2023ICCV,
  title={Hidden Biases of End-to-End Driving Models},
  author={Bernhard Jaeger and Kashyap Chitta and Andreas Geiger},
  booktitle={Proc. of the IEEE International Conf. on Computer Vision (ICCV)},
  year={2023}
}

@article{Yang2023LLM4DriveAS,
  title={LLM4Drive: A Survey of Large Language Models for Autonomous Driving},
  author={Zhenjie Yang and Xiaosong Jia and Hongyang Li and Junchi Yan},
  journal={ArXiv},
  year={2023},
  volume={abs/2311.01043},
}

@article{zhai2023rethinking,
  title={Rethinking the open-loop evaluation of end-to-end autonomous driving in nuscenes},
  author={Zhai, Jiang-Tian and Feng, Ze and Du, Jinhao and Mao, Yongqiang and Liu, Jiang-Jiang and Tan, Zichang and Zhang, Yifu and Ye, Xiaoqing and Wang, Jingdong},
  journal={arXiv preprint arXiv:2305.10430},
  year={2023}
}

@inproceedings{dosovitskiy2017carla,
  title={CARLA: An open urban driving simulator},
  author={Dosovitskiy, Alexey and Ros, German and Codevilla, Felipe and Lopez, Antonio and Koltun, Vladlen},
  booktitle={Conference on robot learning},
  pages={1--16},
  year={2017},
  organization={PMLR}
}

@article{zhai2023ADMLP,
  title={Rethinking the Open-Loop Evaluation of End-to-End Autonomous Driving in nuScenes},
  author={Zhai, Jiang-Tian and Feng, Ze and Du, Jihao and Mao, Yongqiang and Liu, Jiang-Jiang and Tan, Zichang and Zhang, Yifu and Ye, Xiaoqing and Wang, Jingdong},
  journal={arXiv preprint arXiv:2305.10430},
  year={2023}
}

@inproceedings{
jia2025drivetransformer,
title={DriveTransformer: Unified Transformer for Scalable End-to-End Autonomous Driving},
author={Xiaosong Jia and Junqi You and Zhiyuan Zhang and Junchi Yan},
booktitle={The Thirteenth International Conference on Learning Representations},
year={2025},
url={https://openreview.net/forum?id=M42KR4W9P5}
}

@inproceedings{jia2024bench,
  title={Bench2Drive: Towards Multi-Ability Benchmarking of Closed-Loop End-To-End Autonomous Driving},
  author={Xiaosong Jia and Zhenjie Yang and Qifeng Li and Zhiyuan Zhang and Junchi Yan},
  booktitle={NeurIPS 2024 Datasets and Benchmarks Track},
  year={2024}
}

@article{wang2023drivemlm,
  title={Drivemlm: Aligning multi-modal large language models with behavioral planning states for autonomous driving},
  author={Wang, Wenhai and Xie, Jiangwei and Hu, ChuanYang and Zou, Haoming and Fan, Jianan and Tong, Wenwen and Wen, Yang and Wu, Silei and Deng, Hanming and Li, Zhiqi and others},
  journal={arXiv preprint arXiv:2312.09245},
  year={2023}
}

@misc{deepseekai2024deepseekv3technicalreport,
      title={DeepSeek-V3 Technical Report}, 
      author={DeepSeek-AI},
      year={2024},
      eprint={2412.19437},
      archivePrefix={arXiv},
      primaryClass={cs.CL},
      url={https://arxiv.org/abs/2412.19437}, 
}

@inproceedings{renz2025simlingo,
  title={Simlingo: Vision-only closed-loop autonomous driving with language-action alignment},
  author={Renz, Katrin and Chen, Long and Arani, Elahe and Sinavski, Oleg},
  booktitle={Proceedings of the Computer Vision and Pattern Recognition Conference},
  pages={11993--12003},
  year={2025}
}

@article{mei2024continuously,
  title={Continuously learning, adapting, and improving: A dual-process approach to autonomous driving},
  author={Mei, Jianbiao and Ma, Yukai and Yang, Xuemeng and Wen, Licheng and Cai, Xinyu and Li, Xin and Fu, Daocheng and Zhang, Bo and Cai, Pinlong and Dou, Min and others},
  journal={arXiv preprint arXiv:2405.15324},
  year={2024}
}

@article{liu2022rectified,
  title={Rectified flow: A marginal preserving approach to optimal transport},
  author={Liu, Qiang},
  journal={arXiv preprint arXiv:2209.14577},
  year={2022}
}

@inproceedings{hamdan2025eta,
  title={ETA: Efficiency through Thinking Ahead, A Dual Approach to Self-Driving with Large Models},
  author={Hamdan, Shadi and Sima, Chonghao and Yang, Zetong and Li, Hongyang and Guney, Fatma},
  booktitle={Proceedings of the IEEE/CVF International Conference on Computer Vision},
  pages={26529--26538},
  year={2025}
}

@article{fu2025orion,
  title={Orion: A holistic end-to-end autonomous driving framework by vision-language instructed action generation},
  author={Fu, Haoyu and Zhang, Diankun and Zhao, Zongchuang and Cui, Jianfeng and Liang, Dingkang and Zhang, Chong and Zhang, Dingyuan and Xie, Hongwei and Wang, Bing and Bai, Xiang},
  journal={arXiv preprint arXiv:2503.19755},
  year={2025}
}

@inproceedings{yasarla2025dyss,
  title={DySS: Dynamic Queries and State-Space Learning for Efficient 3D Object Detection from Multi-Camera Videos},
  author={Yasarla, Rajeev and Han, Shizhong and Cai, Hong and Porikli, Fatih},
  booktitle={Proceedings of the Computer Vision and Pattern Recognition Conference},
  pages={2510--2519},
  year={2025}
}

@article{jiang2025better,
  title={The better you learn, the smarter you prune: Towards efficient vision-language-action models via differentiable token pruning},
  author={Jiang, Titong and Jiang, Xuefeng and Ma, Yuan and Wen, Xin and Li, Bailin and Zhan, Kun and Jia, Peng and Liu, Yahui and Sun, Sheng and Lang, Xianpeng},
  journal={arXiv preprint arXiv:2509.12594},
  year={2025}
}

@article{liu2026palm,
  title={PALM: Progress-Aware Policy Learning via Affordance Reasoning for Long-Horizon Robotic Manipulation},
  author={Liu, Yuanzhe and Zhu, Jingyuan and Mo, Yuchen and Li, Gen and Cao, Xu and Jin, Jin and Shen, Yifan and Li, Zhengyuan and Yu, Tianjiao and Yuan, Wenzhen and others},
  journal={arXiv preprint arXiv:2601.07060},
  year={2026}
}

@article{xu2025stare,
  title={STARE-VLA: Progressive Stage-Aware Reinforcement for Fine-Tuning Vision-Language-Action Models},
  author={Xu, Feng and Zhai, Guangyao and Kong, Xin and Fu, Tingzhong and Gordon, Daniel FN and An, Xueli and Busam, Benjamin},
  journal={arXiv preprint arXiv:2512.05107},
  year={2025}
}

@article{shen2025fine,
  title={Fine-grained preference optimization improves spatial reasoning in vlms},
  author={Shen, Yifan and Liu, Yuanzhe and Zhu, Jingyuan and Cao, Xu and Zhang, Xiaofeng and He, Yixiao and Ye, Wenming and Rehg, James Matthew and Lourentzou, Ismini},
  journal={arXiv preprint arXiv:2506.21656},
  year={2025}
}

@article{li2026toward,
  title={Toward Cognitive Supersensing in Multimodal Large Language Model},
  author={Li, Boyi and Shen, Yifan and Liu, Yuanzhe and Xu, Yifan and Liu, Jiateng and Li, Xinzhuo and Li, Zhengyuan and Zhu, Jingyuan and Zhong, Yunhan and Lan, Fangzhou and others},
  journal={arXiv preprint arXiv:2602.01541},
  year={2026}
}

@article{su2026drivemamba,
  title={DriveMamba: Task-Centric Scalable State Space Model for Efficient End-to-End Autonomous Driving},
  author={Su, Haisheng and Wu, Wei and Song, Feixiang and Zhang, Junjie and Yang, Zhenjie and Yan, Junchi},
  journal={arXiv preprint arXiv:2602.13301},
  year={2026}
}

@article{yang2025raw2drive,
  title={Raw2drive: Reinforcement learning with aligned world models for end-to-end autonomous driving (in carla v2)},
  author={Yang, Zhenjie and Jia, Xiaosong and Li, Qifeng and Yang, Xue and Yao, Maoqing and Yan, Junchi},
  journal={arXiv preprint arXiv:2505.16394},
  year={2025}
}

@misc{su2026egofsd,
      title={EgoFSD: Ego-Centric Fully Sparse Paradigm with Uncertainty Denoising and Iterative Refinement for Efficient End-to-End Self-Driving}, 
      author={Haisheng Su and Wei Wu and Zhenjie Yang and Isabel Guan},
      year={2026},
      eprint={2409.09777},
      archivePrefix={arXiv},
      primaryClass={cs.CV},
      url={https://arxiv.org/abs/2409.09777}, 
}

@inproceedings{huang2023iddr,
  title={Iddr-ngp: Incorporating detectors for distractors removal with instant neural radiance field},
  author={Huang, Xianliang and Gou, Jiajie and Chen, Shuhang and Zhong, Zhizhou and Guan, Jihong and Zhou, Shuigeng},
  booktitle={Proceedings of the 31st ACM International Conference on Multimedia},
  pages={1343--1351},
  year={2023}
}

@article{huang2026nerf,
  title={NeRF-MIR: Toward High-Quality Restoration of Masked Images With Neural Radiance Fields},
  author={Huang, Xianliang and Zhong, Zhizhou and Chen, Shuhang and Xu, Yi and Guan, Jihong and Zhou, Shuigeng},
  journal={IEEE Transactions on Neural Networks and Learning Systems},
  year={2026},
  publisher={IEEE}
}

@inproceedings{hegde2025distilling,
  title={Distilling multi-modal large language models for autonomous driving},
  author={Hegde, Deepti and Yasarla, Rajeev and Cai, Hong and Han, Shizhong and Bhattacharyya, Apratim and Mahajan, Shweta and Liu, Litian and Garrepalli, Risheek and Patel, Vishal M and Porikli, Fatih},
  booktitle={Proceedings of the Computer Vision and Pattern Recognition Conference},
  pages={27575--27585},
  year={2025}
}

@article{driess2025knowledge,
  title={Knowledge insulating vision-language-action models: Train fast, run fast, generalize better},
  author={Driess, Danny and Springenberg, Jost Tobias and Ichter, Brian and Yu, Lili and Li-Bell, Adrian and Pertsch, Karl and Ren, Allen Z and Walke, Homer and Vuong, Quan and Shi, Lucy Xiaoyang and others},
  journal={arXiv preprint arXiv:2505.23705},
  year={2025}
}

@article{jia2026guidedvla,
  title={GuidedVLA: Specifying Task-Relevant Factors via Plug-and-Play Action Attention Specialization},
  author={Jia, Xiaosong and Yang, Bowen and Ge, Zuhao and Nie, Xian and Zhou, Yuchen and Fan, Cunxin and Li, Yufeng and Chai, Yilin and Jing, Chao and Liang, Zijian and others},
  journal={arXiv preprint arXiv:2605.12369},
  year={2026}
}

@article{jia2026bench2drive,
  title={Bench2Drive-VL: Benchmarks for Closed-Loop Autonomous Driving with Vision-Language Models},
  author={Jia, Xiaosong and Shao, Yuqian and Yang, Zhenjie and Li, Qifeng and Zhang, Zhiyuan and Yan, Junchi},
  journal={arXiv preprint arXiv:2604.01259},
  year={2026}
}

@article{jia2025drivevggt,
  title={DriveVGGT: Visual Geometry Transformer for Autonomous Driving},
  author={Jia, Xiaosong and Liu, Yanhao and You, Junqi and Xia, Renqiu and Hong, Yu and Yan, Junchi},
  journal={arXiv preprint arXiv:2511.22264},
  year={2025}
}

@article{jia2025spatial,
  title={Spatial retrieval augmented autonomous driving},
  author={Jia, Xiaosong and Zhang, Chenhe and Jiang, Yule and Wong, Songbur and Zhang, Zhiyuan and Chen, Chen and Zhang, Shaofeng and Zhou, Xuanhe and Yang, Xue and Yan, Junchi and others},
  journal={arXiv preprint arXiv:2512.06865},
  year={2025}
}

@inproceedings{zhangtrajtok,
  title={TrajTok: What makes for a good trajectory tokenizer in behavior generation?},
  author={Zhang, Zhiyuan and Jia, Xiaosong and Chen, Guanyu and Li, Qifeng and Wu, Zuxuan and Jiang, Yu-Gang and Yan, Junchi},
  booktitle={The Fourteenth International Conference on Learning Representations}
}

@article{han2025percept,
  title={Percept-WAM: Perception-enhanced world-awareness-action model for robust end-to-end autonomous driving},
  author={Han, Jianhua and Tian, Meng and Zhu, Jiangtong and He, Fan and Zhang, Huixin and Guo, Sitong and Zhu, Dechang and Tang, Hao and Xu, Pei and Guo, Yuze and others},
  journal={arXiv preprint arXiv:2511.19221},
  year={2025}
}
